\documentclass{article} 
\usepackage{iclr2026_conference,times}

\usepackage{amsmath,amsfonts,bm}

\def\eqref#1{equation~\ref{#1}}

\def\1{\bm{1}}

\DeclareMathAlphabet{\mathsfit}{\encodingdefault}{\sfdefault}{m}{sl}
\SetMathAlphabet{\mathsfit}{bold}{\encodingdefault}{\sfdefault}{bx}{n}

\usepackage{url}
\usepackage{multirow}
\usepackage[utf8]{inputenc} 
\usepackage[T1]{fontenc}    
\usepackage{url}            
\usepackage{booktabs}       
\usepackage{amsfonts}       
\usepackage{nicefrac}       
\usepackage{microtype}      
\usepackage{xcolor}         
\definecolor{cvprblue}{rgb}{0.21,0.49,0.74}
\def\ours{MR-LoRA}
\definecolor{txblue}{rgb}{0,1,1}
\usepackage[breaklinks=true,colorlinks=true,citecolor=cvprblue,bookmarks=false]{hyperref}
\usepackage{colortbl}  
\usepackage{enumitem}
\usepackage{tcolorbox}
\usepackage{listings}
\usepackage{graphicx}

\usepackage{float}
\usepackage{array}
\newcolumntype{L}{>{\raggedright\arraybackslash}p{1.5cm}}
\usepackage{siunitx}

\usepackage{subcaption}
\usepackage{cleveref}  
\usepackage{wrapfig}
\definecolor{Light}{HTML}{e7f5fe}
\newcommand{\MLLMCLemoji}{\includegraphics[height=1.4\fontcharht\font`\B]{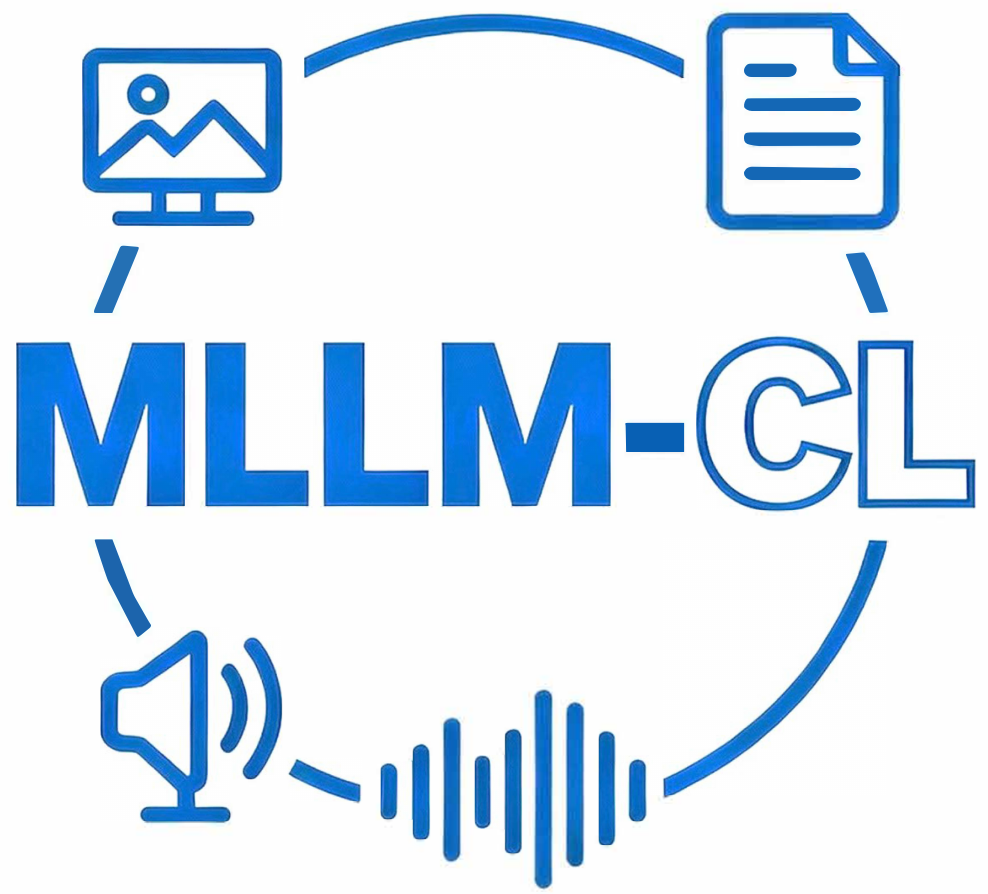}}

\title{\centering \MLLMCLemoji{} MLLM-CL: Continual Learning for Multimodal Large Language Models}
\usepackage{mathrsfs}
\newlength\savewidth

\Crefname{figure}{Fig.}{Figs.}
\Crefname{section}{Sec.}{Secs.}
\Crefname{table}{Tab.}{Tabs.}

\Crefname{appendix}{\textit{Appendix}}{\textit{Appendices}}

\author{
\begin{minipage}[t]{\textwidth}
\centering\hspace*{-2em}
Hongbo Zhao$^{1,2}$ \quad Fei Zhu$^{3}$  \quad Haiyang Guo$^{1,2}$  \quad Meng Wang$^3$    \\[0.1cm]
\centering\hspace*{-2em}
Rundong Wang$^{3,4}$ \quad Gaofeng Meng$^{1,2,3}$  \quad Zhaoxiang Zhang$^{1,2}$\\[0.3cm]
$^1$UCAS \quad $^2$CASIA \quad  $^3$HKISI, CAS \quad $^4$HKU \\[0.3cm]
\makebox[0.7\textwidth][l]{\includegraphics[height=1.0em]{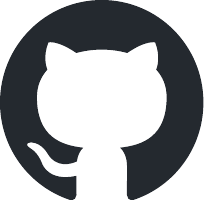}\qquad~\url{https://github.com/bjzhb666/MLLM-CL}} \\
\makebox[0.7\textwidth][l]{\includegraphics[height=1.0em]{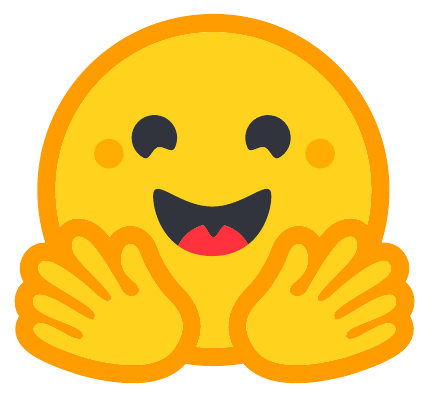}\qquad~\url{https://huggingface.co/MLLM-CL}}\\
\makebox[0.7\textwidth][l]{\includegraphics[height=1.0em]{logo/hf-logo.pdf}\qquad~\url{https://huggingface.co/datasets/MLLM-CL/MLLM-CL}} \\
\makebox[0.7\textwidth][l]{\includegraphics[height=0.8em]{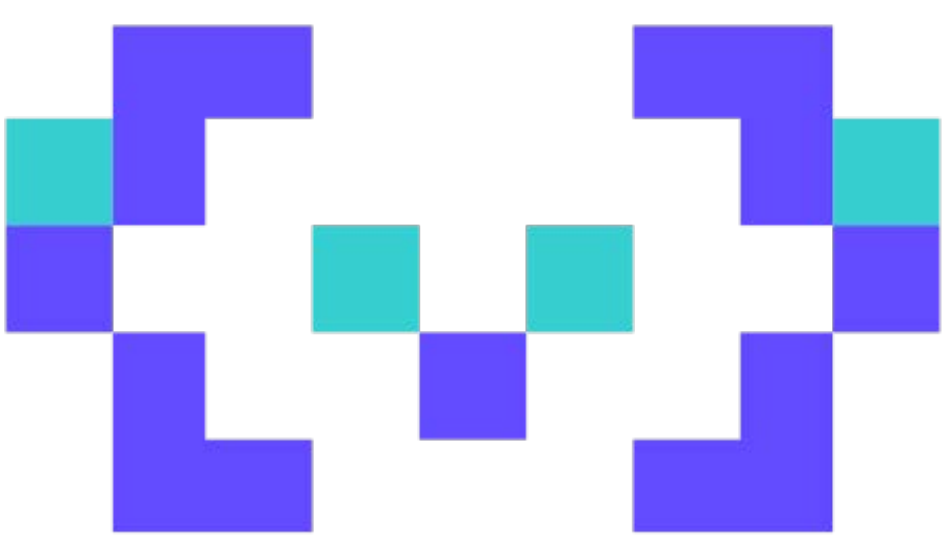}\quad \ \; ~\url{https://modelscope.cn/organization/MLLM-CL}} \\
\makebox[0.7\textwidth][l]{\includegraphics[height=0.8em]{logo/modelscope-logo.pdf}\quad \ \; ~\url{https://modelscope.cn/datasets/MLLM-CL/MLLM-CL}} \\
\end{minipage}
}

\iclrfinalcopy 
\begin{document}

\maketitle
\begin{figure}[h]
    \centering
    \includegraphics[width=1.0\linewidth]{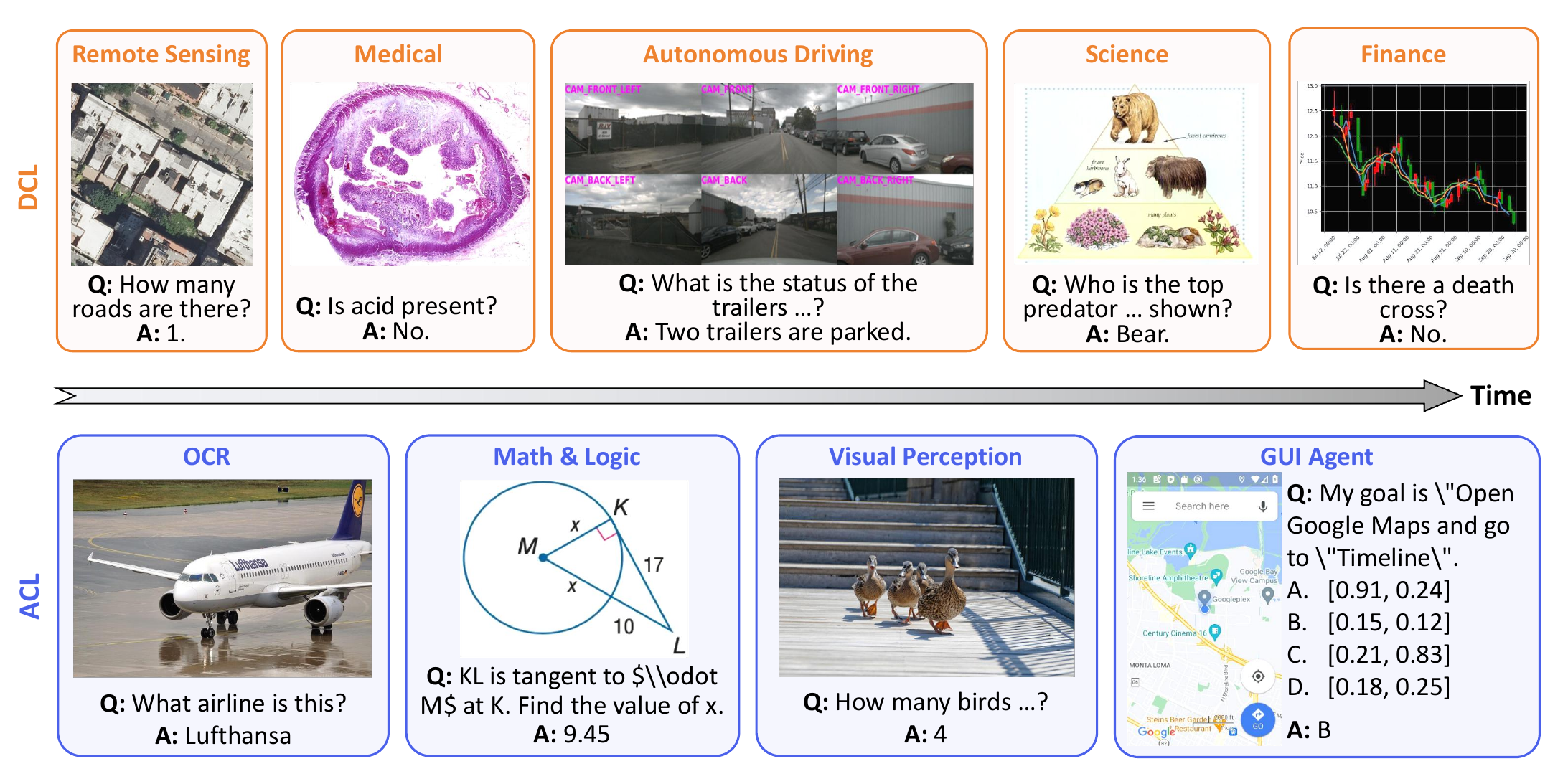}
    \caption{Demonstrations of MLLM-CL benchmark.
    It incorporates Domain Continual Learning~(DCL), which adds domain-specific knowledge, and Ability Continual Learning (ACL), which improves fundamental abilities for multimodal large language models.}
    \label{fig:benchmark}
\end{figure}
\begin{abstract}
  Recent Multimodal Large Language Models (MLLMs) excel in vision-language understanding but face challenges in adapting to dynamic real-world scenarios that require continuous integration of new knowledge and skills. While continual learning (CL) offers a potential solution, existing benchmarks and methods suffer from critical limitations. In this paper, we introduce MLLM-CL, a novel benchmark encompassing domain and ability continual learning, where the former focuses on independently and identically distributed (IID) evaluation across evolving mainstream domains, whereas the latter evaluates on non-IID scenarios with new model abilities. Methodologically, we propose preventing catastrophic interference through parameter isolation and an MLLM-based routing mechanism. Extensive experiments demonstrate that our approach can integrate domain-specific knowledge and functional abilities with minimal forgetting, significantly outperforming existing methods. Our benchmark and code are available at \url{https://github.com/bjzhb666/MLLM-CL}.
\end{abstract}

\section{Introduction}

Recent advancements in Multimodal Large Language Models (MLLMs) \citep{liu2023improvedllava, chen2024allava} have demonstrated remarkable capabilities in vision-language understanding. These models typically undergo supervised finetuning on carefully curated multi-task datasets, whereas real-world applications require continuous adaptation to evolving user requirements and dynamic data streams with shifting domain distributions. To incorporate new knowledge and skills, full retraining of large models is costly in both time and computing resources; besides, straightforward finetuning on novel tasks often results in catastrophic forgetting \citep{mccloskey1989catastrophic, zhai2023investigating}. Therefore, for deployment in ever-changing environments, there is an urgent need to develop MLLMs capable of continually consolidating new skills while maintaining performance on prior tasks.
Recently, a few studies \citep{chen2024coin,zeng2024modalprompt,cao2024continual,guo2025hide,he2023continual} have explored continual learning (CL) of MLLMs. However, current works still have key limitations in both benchmarks and methodologies, preventing them from effectively exploring CL in MLLMs. 

Firstly, there is a lack of well-established benchmarks. \citet{chen2024coin} proposed the first continual instruction tuning benchmark for MLLMs comprising several downstream datasets, while some of them have already been learned during the early supervised finetuning (SFT) phase of MLLM. 
\citet{huai2025cl} divided VQAv2 \citep{goyal2017making} into several tasks and conducted continual instruction tuning directly from the LLaVA~\citep{liu2023llava} base model. However, in real-world applications, continually learning subsets of a specific dataset is impractical, and it is unlikely to finetune an MLLM on downstream tasks without any SFT on general multimodal data. Moreover, those benchmarks only consider independently and identically distributed (IID) evaluation (the training and test sets are split from the same dataset), while the model would encounter non-IID inputs in practice.

Secondly, existing methods have notable limitations: (1) Some approaches share the same set of parameters for different tasks \citep{chen2024coin, huang2024class}. This might be suitable for a conventional class-incremental learning scenario where different tasks often belong to the same dataset. However, MLLMs often encounter inputs from various domains, and the inherent task conflicts \citep{wei2025asymlora, yang2024low} would lead to loss of plasticity during continual learning, particularly when handling heterogeneous modalities across divergent domains. (2) Parameter isolation methods have to determine which task-specific parameters to apply for a given input during inference. This selection is usually driven by simple hand-crafted similarity metrics \citep{zeng2024modalprompt, guo2025hide}, which can be unreliable when confronted with complex multimodal data, consequently undermining overall performance.

In this paper, we establish a novel benchmark MLLM-CL, which includes two practical settings, \textit{i.e.}, domain continual learning (DCL) and ability continual learning (ACL), as shown in Fig. \ref{fig:benchmark}. Specifically, DCL aims to equip the model with domain-specific knowledge continually by learning and evaluating on several mainstream domains (remote sensing, medical, autonomous driving, science, and finance), where the training and test sets are IID. Differently, ACL focuses on incorporating fundamental abilities (OCR, math \& logic, visual perception, and GUI agent), which are evaluated on non-IID test sets. Together, these two settings provide a comprehensive and realistic evaluation for continual learning of MLLMs. 

Further, we design a novel method to build an efficient, lifelong-evolving MLLM. For plasticity preservation, we employ domain or ability-specific Low-Rank Adaptation (LoRA) modules~\citep{hu2021lora} that maintain parameter isolation across sequentially arriving tasks, enabling comprehensive acquisition of new knowledge while preventing catastrophic interference through explicit architectural decoupling. Concurrently, to enhance parameter selection accuracy in complex multimodal scenarios, we devise a multimodal routing mechanism that leverages the model's intrinsic multimodal understanding capabilities to automatically align input patterns with optimal task parameters. This strategy effectively transforms the MLLM’s knowledge into an explicit expert selector. 

In summary, our main contributions are as follows:
\begin{itemize} 
	\item We establish a novel benchmark for CL of MLLMs, with practical domain and ability continual learning settings, focusing on both IID and non-IID evaluation.
    \item We propose a simple yet effective method with domain or ability-specific low-rank adaptation and large multimodal model-based parameter selection. 
    \item Experiments show that our method achieves impressive results on both domain and ability settings of the MLLM-CL benchmark, significantly outperforming existing approaches. 
\end{itemize}
\section{Related Work}
\paragraph{Continual Learning.}
Researchers have developed primarily four main strategies for continual learning: rehearsal-based methods~\citep{lavda2018continual,buzzega2020dark}, regularization-based methods \citep{kirkpatrick2017overcoming, li2017learning}, structure-based methods \citep{mallya2018piggyback, Douillard_Rame_Couairon_Cord_2022}, and prompt-based methods \citep{l2p, smith2023coda}.
CL in large language models has recently gained much attention \citep{wu2024continual,shi2024continual}.
According to the training stages, we can divide them into continual pre-training \citep{jang2022temporalwiki, cossu2024continual}, continual instruction tuning~\citep{razdaibiedina2023progressive,zan2022cert,yin2022contintin,wang2023orthogonal}, and continual alignment~\citep{zhang2024cppo,suhr2024continual}.
However, few studies focus on continual learning of MLLMs \citep{chen2024coin,zeng2024modalprompt,cao2024continual,guo2025hide,guo2025comprehensive}.
These prior attempts establish benchmarks with a simple \textit{dataset incremental setting} where training and test sets are distributed independently and identically.
Some works focus on conducting continuous instruction tuning directly from the model after the pretraining process~\citep{huai2025cl,he2023continual}.
While these efforts have advanced the development of continual learning for MLLMs to some extent, they exhibit an apparent gap with the real-world production environment. Therefore, our work fills this gap and proposes a comprehensive and practical benchmark, including adding domain-specific knowledge and general abilities for CL of MLLM.

\paragraph{Multimodal Large Language Models.}
Recent advances in MLLMs have demonstrated remarkable capabilities in multimodal understanding, open-ended generation, and instruction following across modalities.
Early efforts, such as LLaVA \citep{liu2023llava,liu2023improvedllava} and Qwen-VL \citep{bai2023qwen}, use image encoders \citep{radford2021learning} and projectors to transfer multimodal inputs into language embedding space.
Recent advances~\citep{gpt4o, li2024llava, Qwen2.5-VL, fu2025vita} expand the ability of MLLM into more modalities, such as video and audio.
With the rapid growth of MLLMs, the costs associated with training from scratch have increased dramatically \citep{li2024llava, tong2024cambrian, Qwen2.5-VL,chen2024expanding}.
Therefore, adapting MLLMs to dynamic environments by retraining them from scratch becomes expensive and inefficient, creating an imperative demand for continual learning of MLLMs.


\section{MLLM-CL Benchmark}
In this section, we provide the problem formulation and introduce the continual learning benchmark MLLM-CL. 
Based on the general ability and domain-specific knowledge updated in the instruction tuning stage, we divide our benchmark into domain continual learning and ability continual learning, respectively.
In domain continual learning, we desire the model to learn knowledge continually, and the training sets and the test sets are IID.
While in ability continual learning, we desire the model to enhance different abilities from the training data and generalize to non-IID test sets.

\textbf{Problem Statement.}
Continual learning in MLLMs involves sequentially learning a series of multimodal tasks. Let $\mathcal{X}^{\text{img}}$ and $\mathcal{X}^{\text{ins}}$ denote the image and instruction spaces, respectively, and $\mathcal{Y}$ represent the label space for answers composed of $L$ tokens. Given a sequence of datasets $\mathcal{D}_1, \ldots, \mathcal{D}_T$, where each $\mathcal{D}_t = \{(x^{\text{img}}_{t,i}, x^{\text{ins}}_{t,i}, y_{t,i})\}_{i=1}^{N_t}$ contains $N_t$ image-instruction-answer triplets drawn IID from the task-specific distribution $\mathcal{P}_t = \mathcal{X}_t^{\text{img}} \times \mathcal{X}_t^{\text{ins}} \times \mathcal{Y}_t$. Our goal is to continually update a multimodal model on observed data while retaining knowledge from previous tasks. Denote the model by \(f\) with parameters \(\theta_t\) at stage $t$, the training objective of MLLM is to predict the next token in an autoregressive way:
\begin{equation}\label{eq1}
\begin{aligned}
\mathcal{L}_{\mathrm{MLLM}}(\theta_t)
= - \sum_{i=1}^{N_t} \sum_{l=1}^{L}
\log p_{\theta_t}(y_{t,i}^{l}|x^{\text{img}}_{t,i}, x^{\text{ins}}_{t,i}, y_{t,i}^{<l}).
\end{aligned}
\end{equation}
At inference time, given an image-instruction pair $(x^{\text{img}}, x^{\text{ins}})$ drawn from all learned task distributions $\{\mathcal{P}_j\}_{j=1}^{t}$, the model generates tokens autoregressively, \textit{i.e.}, the $l$-th output token is $\hat{y}^l = \mathop{\arg\max}\limits_{v \in \mathcal{V}} ~p_{\theta}(v|x^{\text{img}}, x^{\text{text}}, \hat{y}^{<l})$. 
The above describes a typical IID scenario (\textit{e.g.}, domain-specific evaluation) where training and test data belong to $\{\mathcal{P}_j\}_{j=1}^{t}$. In practice, the model can encounter various out-of-distribution inputs $\{\mathcal{P}_{j,\text{non-iid}}\}_{j=1}^{t} \neq \{\mathcal{P}_{j}\}_{j=1}^{t}$ (\textit{e.g.}, ability evaluation where the input images and instruction style can be diverse), and the model is supposed to handle such a non-IID scenario.
\begin{table}[t]
    \centering\small
    \caption{Statistics of the training datasets and test datasets for domain continual learning and ability continual learning. 
    In domain continual learning, "RS" stands for remote sensing, "Med" is medical, "AD" is autonomous driving, "Sci" stands for science, and "Fin" means finance.
    In ability continual learning, "M \& L" stands for math \& logic. "VP" means visual perception.}
    \vskip -0.1in
    \renewcommand{\arraystretch}{1}
    \scalebox{0.9}{
    \begin{tabular}{lllcc}
    \toprule
Task & Train Dataset & Test Dataset & Train Number & Test Number  \\ \midrule 
\multicolumn{5}{c}{{Domain Continual Learning}} \\ \midrule
      RS  &  RSVQA & RSVQA & 60k & 10k \\
      Med  & PathVQA & PathVQA & 22.8k & 9.8k\\
      AD & DriveLM & DriveLM & 60k & 10k\\
      Sci &\begin{tabular}[c]{@{}l@{}}AI2D, SciVerse \\ MapQA, TQA\end{tabular}  & \begin{tabular}[c]{@{}l@{}}AI2D, SciVerse \\ MapQA, TQA\end{tabular} & \begin{tabular}[c]{@{}c@{}} 33.4k\\ (12.4k, 0.9k, 9.6k, 7.8k)\end{tabular}  & \begin{tabular}[c]{@{}c@{}} 8.2k \\ (3.1k, 0.2k, 2.4k, 1.9k)\end{tabular} \\
      Fin & StockQA &StockQA & 60k & 10k\\ \midrule
      \multicolumn{5}{c}{{Ability Continual Learning}} \\ \midrule
      OCR & Monkey & OCRBench &128.1k & 1k \\
      M \& L & MathV360K, MAVIS & MathVista & 526.1k & 1k\\
      VP & CLEVR, TallyQA & CV-Bench &119.9k &0.8k\\
      GUI Agent & \begin{tabular}[c]{@{}l@{}}ScreenQA, MultiUI \\Screen2Words\end{tabular} & MMTBench &147.3k & 0.8k\\
      \bottomrule
    \end{tabular}
    }
    \label{Tab:static}
    \vskip -0.1in
\end{table}

\textbf{Domain Continual Learning (DCL).} 
Continually adding domain knowledge is crucial for constructing a powerful MLLM.
To achieve this goal, we propose domain continual learning and choose five mainstream and common domains: remote sensing, medical, science, autonomous driving, and finance.
Specifically, we choose RSVQA~\citep{lobry2020rsvqa}, PathVQA~\citep{he2020pathvqa}, DriveLM~\citep{sima2023drivelm}, FinVis~\citep{wang2023finvis}, AI2D~\citep{kembhavi2016diagram}, SciVerse~\citep{guo2025sciverse}, MapQA~\citep{chang2022mapqa} and TQA~\citep{kembhavi2017you}.
However, FinVis is a
\begin{wrapfigure}{r}{7.8cm}
\vskip -0.1in
\includegraphics[width=7.8cm]{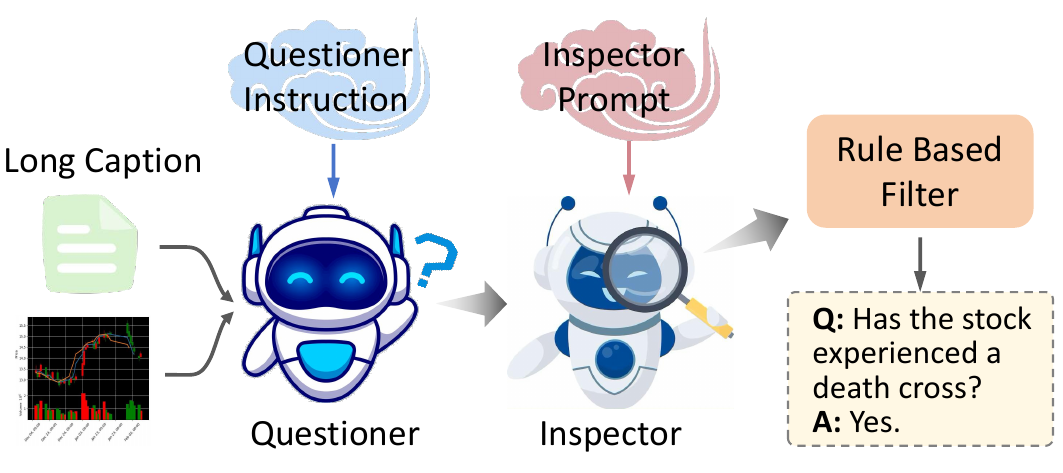}
\caption{The questioner-inspector data pipeline for generating StockQA instruction tuning dataset.}
\vskip -0.1in
\label{Fig:data-pipeline}
\end{wrapfigure}
caption dataset in Chinese, which may result in a language gap and is not convenient for evaluation.
Therefore, we regenerate the SFT and test data as multi-choice questions and yes-or-no questions using a \textit{questioner-inspector} data pipeline.
\Cref{Fig:data-pipeline} shows the overall data pipeline.
We use two agents, a QA generator and an inspector.
Considering the varying task difficulties, we use Qwen2.5-VL-72b \citep{Qwen2.5-VL} to generate multiple choice QA pairs and Qwen2.5-VL-7b to generate Y/N QA pairs. 
For the inspector, we use Qwen2.5-VL-7b to check the correctness of each QA pair.
After initial inspection, rule-based formatting is applied to generate the final dataset, named StockQA.
All experiments are conducted using the vllm~\citep{kwon2023efficient} engine.
\Cref{appdx:stockqa} provides detailed prompts for each agent, rules for filtering, examples, and statistics of the StockQA dataset.
\Cref{Tab:static} shows the statistics of the datasets for DCL and \Cref{fig:benchmark} shows some examples.
More examples are provided in the \Cref{appdx:bench:sample}.

\textbf{Ability Continual Learning (ACL).} 
DCL assumes that training and test data are IID.
However, achieving IID between training and test sets is often challenging in real-world scenarios, which has been ignored by 
existing benchmarks~\citep{chen2024coin,zeng2024modalprompt,guo2025hide,cao2024continual}.
Therefore, we consider a more challenging setting with non-IID training and test data, which we term ability continual learning.
For ACL, we select four fundamental abilities for the MLLM to learn sequentially: OCR, math \& logic, visual perception, and GUI agent.
In terms of the SFT data, we collect the training data from LLaVA-OneVision~\citep{li2024llava}, Monkey~\citep{li2023monkey}, ScreenQA~\citep{hsiao2022screenqa}, Screen2Words~\citep{wang2021screen2words}, MultiUI~\citep{liu2024harnessing}, Math-LLaVA~\citep{shihu2024mathllava}, MAVIS~\citep{zhang2024mavismathematicalvisualinstruction}, CLVER~\citep{johnson2017clevr} and TallyQA~\citep{acharya2019tallyqa} and testing data from OCRBench~\citep{OCRBench}, MathVista~\citep{lu2024mathvista}, MMTBench-GUI~\citep{mmtbench} and CV-Bench-Counting~\citep{tong2024cambrian}, respectively.
\Cref{Tab:static} presents the details of the datasets for training and testing in ACL, and \Cref{fig:benchmark} provides a demonstration.
Additional examples can be found in the \Cref{appdx:bench:sample}.

\section{The Proposed Method:~\ours}

\subsection{Training: Expert Learning without Task Conflict} \label{sec:conflict}

\begin{figure}[t]
    \centering
    \includegraphics[width=\linewidth]{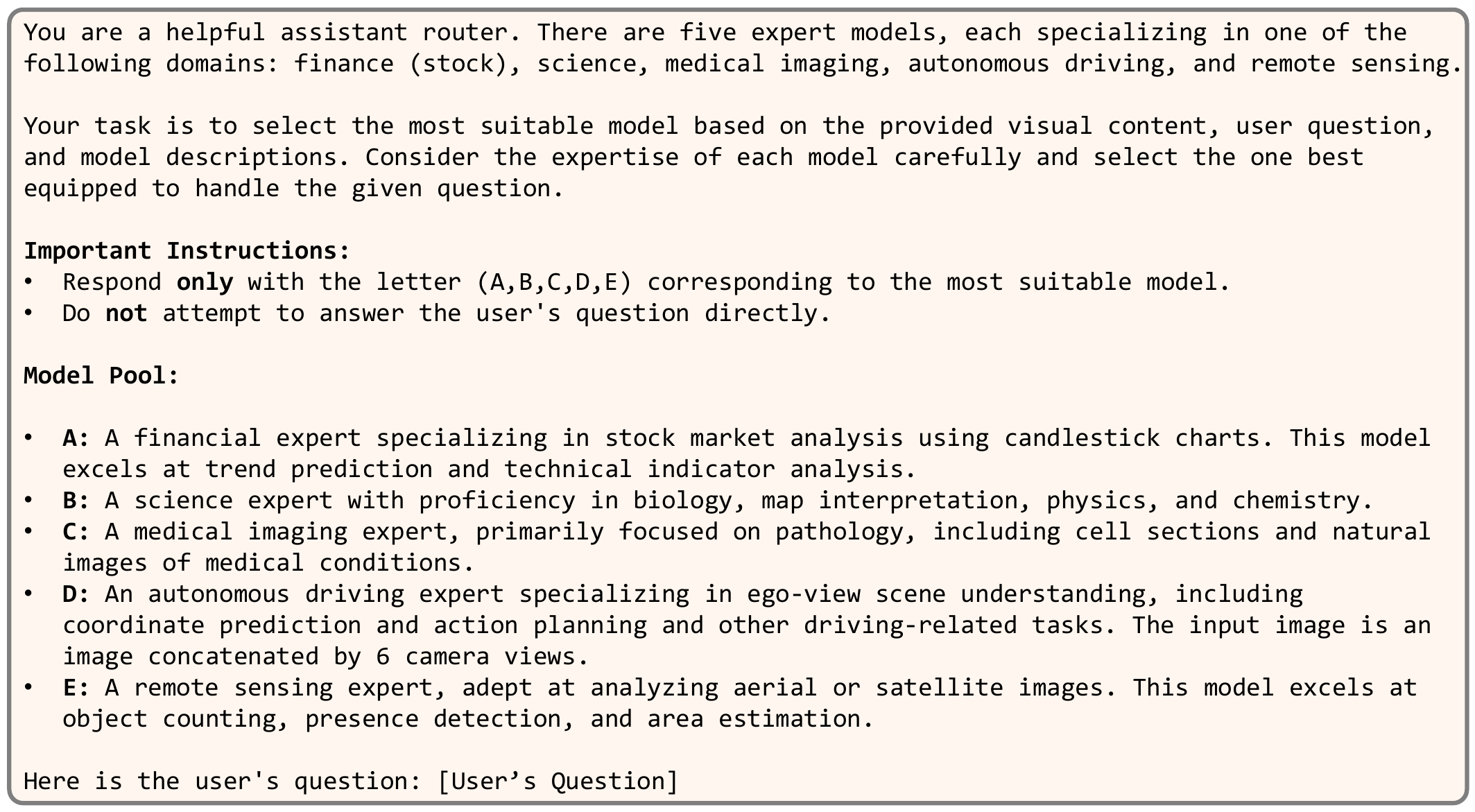}
    \vskip -0.05in
    \caption{Prompt of the MLLM-based router selector.}
    \label{fig:router_instruct}
    \vskip -0.1in
\end{figure}

\begin{wrapfigure}{r}{7.5cm}
\vskip -0.03in
\includegraphics[width=7.5cm]{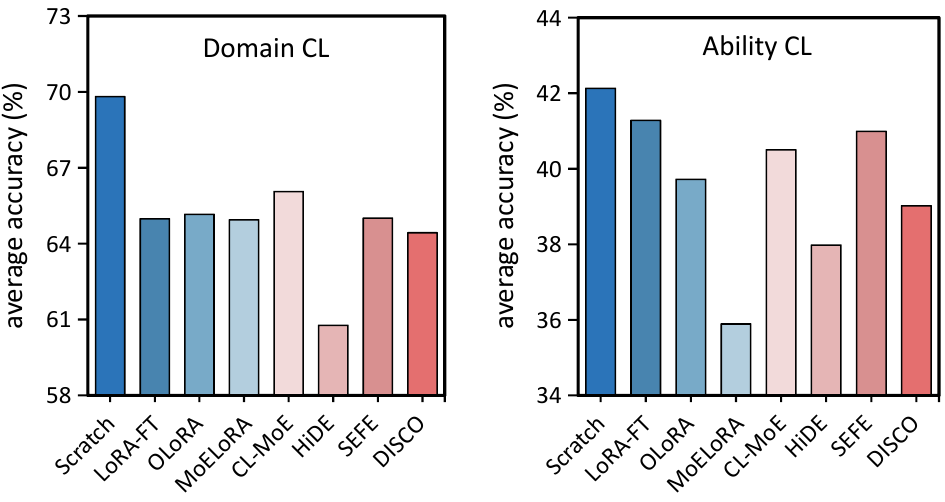}
\vskip -0.1in
\caption{Comparison of new task performance (LLaVA-based) on both domain and ability CL. }
\vskip -0.15in
\label{fig:conflict}
\end{wrapfigure}
\textbf{Learning Low-Rank Expert without Task Conflict.}
In traditional continual learning, particularly class-incremental learning, the model for learning a new task is typically initialized with parameters from the previous task to facilitate knowledge transfer, and then various regularization constraints are incorporated to mitigate catastrophic forgetting. Therefore, a natural question arises: Is this paradigm suitable for continual learning in MLLMs? Some studies \citep{wei2025asymlora, yang2024low} have revealed that data interference widely exists in the training of MLLMs. 
We empirically investigate the task conflict problem of domain and ability continual learning by comparing the average new task performance. 
The results in \Cref{fig:conflict} yield the following observation: (1) Initializing with weights from prior tasks (\emph{e.g.}, LoRA-FT, MoELoRA~\citep{chen2024coin}) reduces model plasticity, leading to worse performance than learning each task individually with randomly initialized LoRA (\emph{i.e.}, scratch). (2) Regularization (\emph{e.g.}, O-LoRA~\citep{wang2023orthogonal}, SEFE~\citep{chen2025sefe}) or parameter-sharing-based methods (\emph{e.g.}, CL-MoE~\citep{huai2025cl}, HiDE~\citep{guo2025hide}) also suffer from loss of plasticity when learning new tasks. (3)~The task conflict in DCL is more severe than that in ACL, which is reasonable because the domain gap in DCL (\emph{e.g.}, autonomous driving vs. science) is often larger than that in ACL (OCR vs. Math).
Based on the above analysis, we propose initializing a fresh LoRA~\citep{hu2021lora} module from scratch for each task to circumvent inter-task conflicts when learning new domains. Compared to the original parameters of the large model, LoRA introduces minimal additional parameters, enabling domain-specific adaptation via lightweight, task-exclusive adapters.

\textbf{Few-shot Router Tuning.}
In our framework, we tune a low-rank expert for each domain or capability, and dynamically select the most appropriate expert at inference time. While existing selection strategies \citep{zeng2024modalprompt, guo2025hide} rely on simple similarity measures, \textit{e.g.}, computing cosine similarity between task prototypes and sample features in the embedding space, multimodal scenarios involve more complex inputs. Therefore, we propose leveraging the MLLM's intrinsic capability to process complex multimodal inputs by tuning an MLLM-based selection router. This router identifies the corresponding expert for each input. Specifically, for each task, we collect a few-shot set $\mathcal{M}_t = \{(x^{\text{img}}_{t,i}, x^{\text{ins}}_{t,i})\}_{i=1}^{m}$, where $m \ll N_t$ (we set $m=20$ in all experiments). After each continual learning phase, the accumulated few-shot data $\{\mathcal{M}_j\}_{j=1}^t$ and expert model descriptions are transformed into structured instructions.
We adopt a \textit{generative} style to select the most suitable expert and tune the MLLM using a router LoRA via autoregressive loss \citep{liu2023improvedllava}. 
An illustration of the router selection prompt for domain continual learning is provided in \Cref{fig:router_instruct}.
\subsection{Inference: Router Selection with MLLM}

\begin{figure}[t]
    \centering
    \vskip -0.1in
    \includegraphics[width=1.0\linewidth]{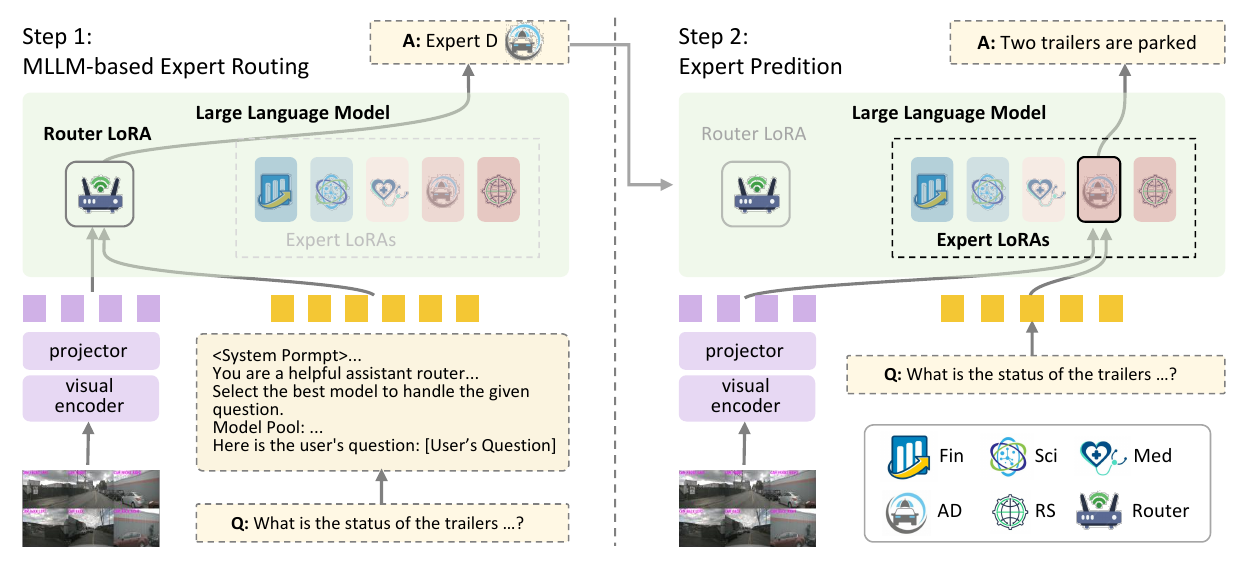}
    \vskip -0.1in
    \caption{Overall framework of our MR-LoRA.}
    \vskip -0.1in
    \label{fig:framework}
\end{figure}
\textbf{Framework of~\ours.}
During inference, with expert learning and router selection, the overall framework of the proposed method is illustrated in \Cref{fig:framework}. Our MR-LoRA performs two-stage inference for a given multimodal input, consisting of a routing phase followed by a prediction phase. In the first stage, the expert selection router is performed to select a domain or ability-specific expert. Then, the selected expert is combined with the pre-trained backbone to output the final response. On the one hand, by decoupling the learning of different domains or abilities, we avoid potential distribution conflict and can learn a good expert for a given task. On the other hand, the proposed router selection strategy largely explores the advantages of MLLMs to improve the flexibility and accuracy of expert selection, ensuring promising final prediction performance during continual learning.
The proposed MLLM-based routing mechanism offers notable advantages: (1) The MLLM's strong multimodal understanding capacity ensures robust expert selection performance on complex multimodal inputs. (2) The selection router is parameter-efficient and learned with few-shot unlabeled image-question pairs, allowing on-the-fly adaptation. 

\section{Experiments}
\subsection{Experimental Setup}
\textbf{Model and Compared Methods.} We conduct experiments on LLaVA-v1.5-7b~\citep{liu2023llava} and InternVL~\citep{chen2024internvl} to continually increase the domain-specific knowledge and abilities in our MLLM-CL benchmark, respectively.
All the continual learning experiments start from the instruct models, \textit{i.e.}, LLaVA-v1.5-7b and InternVL-Chat-V1.0.
For the task sequence in domain continual learning, we choose a random order of remote sensing$\to$medical$\to$autonomous driving$\to$science$\to$finance.
For ability continual learning, we set the task sequence as OCR$\to$math \& logic$\to$visual perception$\to$GUI agent.
We choose CL-MoE~\citep{huai2025cl}, SEFE~\citep{chen2025sefe}, DISCO~\citep{guo2025federated}, O-LoRA~\citep{wang2023orthogonal}, HiDE~\citep{guo2025hide}, MoELoRA~\citep{chen2024coin}, and LoRA~\citep{hu2021lora} as baselines using the MCITlib~\citep{guo2025mcitlib} to show the effectiveness of our proposed method in the two settings of MLLM-CL.
We also report the zero-shot and oracle performance for each setting.
Oracle performance is achieved by training an individual LoRA from the base model and subsequently evaluating its performance.

\textbf{Evaluation Metric.}
We report the last accuracy, which is the accuracy of all seen tasks after learning the last task, mean finetune accuracy (MFT), mean final accuracy (MFN), mean average accuracy (MAA), and backward transfer (BWT) following standard metrics in continual learning~\citep{guo2025hide,chen2025sefe}. 
The detailed calculation of each metric is shown in the \Cref{appdx:eval:metrics}.

\begin{table}[t]
\centering
\small
\setlength{\tabcolsep}{7.3pt}
\caption{Results for LLaVA-based domain continual learning in MLLM-CL benchmark. $^{*}$ denotes
the original method with replay data.} 
\vskip -0.1in
\renewcommand{\arraystretch}{1.05}
\scalebox{0.85}{
\begin{tabular}{l|ccccc|cccc}
\toprule Method & RS    & Med   & AD    & Sci   & Fin   & MFT$\uparrow$ & MFN$\uparrow$ & MAA$\uparrow$ & BWT$\uparrow$ \\
\midrule 
Zeroshot                                     & 32.29 & 28.28 & 15.59 & 35.55 & 62.56 & 34.85 & -     & -     & -      \\
Oracle                                       & 81.06 & 65.83 & 54.17 & 56.86 & 91.14 & 69.81 & -     & -     & -      \\
\midrule \midrule 
LoRA-FT~\citep{hu2021lora} & 69.65 & 41.59 & 25.43 & 40.88 & 87.45 & 64.98 & 53.00 & 61.13 & -14.97 \\
LoRA-FT$^*$~\citep{hu2021lora} & 76.54 & 50.27 & 43.01 & 43.32 & 89.85 & 66.32 & 60.60 & 64.72 & -7.15 \\
O-LoRA~\citep{wang2023orthogonal} & 74.64 & 44.42 & 30.02 & 41.47 & 87.15 & 65.16 & 55.54 & 62.12 & -12.03 \\
O-LoRA$^*$~\citep{wang2023orthogonal} & 76.94 & 41.17 & 34.18 & 39.61 & 83.22 & 60.49 & 55.02 & 60.73 & -6.83 \\
MoELoRA~\citep{chen2024coin} & 77.54 & 41.85 & 27.62 & 40.13 & 86.75 & 64.94 & 54.78 & 61.76 & -12.70 \\
MoELoRA$^*$~\citep{chen2024coin} & 77.63 & 49.54 & 39.08 & 41.04 & 89.21 & 66.24 & 59.30 & 64.81 & -8.68 \\
CL-MoE~\citep{huai2025cl} & 71.34 & 46.84 & 26.33 & 41.17 & 88.74 & 66.06 & 54.88 & 61.79 & -13.96 \\
CL-MoE$^*$~\citep{huai2025cl} & 76.58 & 52.31 & 39.65 & 45.64 & 90.21 & 66.65 & 60.88 & 64.95 & -7.22 \\
HiDe~\citep{guo2025hide} & 74.31 & 48.95 & 33.21 & 38.54 & 81.55 & 60.77 & 55.31 & 60.68 & -6.82 \\
HiDe$^*$~\citep{guo2025hide} & 74.80 & 42.29 & 34.03 & 38.01 & 79.22 & 60.83 & 53.67 & 61.81 & -8.95 \\
SEFE~\citep{chen2025sefe} & 77.26 & 50.37 & 37.21 & 40.87 & 86.82 & 65.01 & 58.51 & 63.63 & -8.13 \\
SEFE$^*$~\citep{chen2025sefe} & 78.43 & 52.85 & 46.21 & 47.76 & 89.33 & 66.89 & 62.92 & 66.51 & -4.97 \\
DISCO~\citep{guo2025federated} & 76.03 & 45.20 & 43.79 & 42.33 & 88.95 & 64.43 & 59.26 & 63.35 & -6.46 \\
DISCO$^*$~\citep{guo2025federated} & 77.78 & 46.25 & 50.45 & 49.51 & 89.71 & 65.27 & 62.74 & 64.92 & -3.17 \\
\rowcolor{Light}MR-LoRA~(Ours) & \textbf{80.87} & \textbf{65.32} & \textbf{54.12} & \textbf{56.71} & \textbf{91.12} & \textbf{69.64} & \textbf{69.63} & \textbf{71.06} & \textbf{-0.01} \\
\bottomrule
\end{tabular}
}
\label{tab:llavaDCL}
\end{table}

\begin{table}[t]
\centering \small
\setlength{\tabcolsep}{8pt}
\caption{Results for LLaVA-based ability continual learning in MLLM-CL benchmark.}
\vskip -0.1in
\renewcommand{\arraystretch}{1.05}
\scalebox{0.85}{
\begin{tabular}{l|cccc|cccc}
\toprule
Method & OCR   & M\&L  & VP    & GUI Agent & MFT$\uparrow$ & MFN$\uparrow$ & MAA$\uparrow$ & BWT$\uparrow$ \\
\midrule
Zeroshot                                & 31.20 & 30.20 & 60.79 & 10.00 & 33.05 &  -   &   -   &  -   \\
Oracle                                  & 33.60 & 36.50 & 65.10 & 32.50 & 41.93 &   -   &   -   &  -   \\
\midrule \midrule
LoRA-FT~\citep{hu2021lora} & 23.60 & 33.70 & 55.84 & 32.50 & 41.28 & 36.41 & 36.58 & -6.49 \\
LoRA-FT$^*$~\citep{hu2021lora} & 21.80 & 32.70 & 58.38 & 28.75 & 40.32 & 35.41 & 36.32 & -6.55 \\
O-LoRA~\citep{wang2023orthogonal} & 29.60 & 32.90 & 52.41 & \textbf{33.75} & 39.72 & 37.16 & 35.42 & -3.41 \\
O-LoRA$^*$~\citep{wang2023orthogonal} & 29.60 & 31.30 & 60.79 & 27.50 & 39.96 & 37.30 & 36.34 & -3.55 \\
MoELoRA~\citep{chen2024coin} & 26.70 & 32.80 & 56.85 & 27.22 & 39.45 & 35.89 & 36.07 & -4.75 \\
MoELoRA$^*$~\citep{chen2024coin} & 19.80 & 32.20 & 54.19 & 30.00 & 40.35 & 34.05 & 35.39 & -8.41 \\
CL-MoE~\citep{huai2025cl} & 19.90 & 32.70 & 53.43 & 30.69 & 40.50 & 34.18 & 35.65 & -8.43 \\
CL-MoE$^*$~\citep{huai2025cl} & 25.40 & 31.80 & 60.91 & 30.00 & 41.22 & 37.03 & 37.28 & -5.59 \\
HiDe~\citep{guo2025hide} & 24.60 & 32.10 & 46.32 & 28.75 & 37.98 & 32.94 & 34.60 & -6.72 \\
HiDe$^*$~\citep{guo2025hide} & 24.60 & 28.40 & 30.71 & 23.75 & 36.84 & 26.86 & 33.54 & -13.30 \\
SEFE~\citep{chen2025sefe} & 26.00 & 33.40 & 57.74 & \textbf{33.75} & 40.98 & 37.72 & 36.59 & -4.35 \\
SEFE$^*$~\citep{chen2025sefe} & 25.60 & 34.80 & 57.61 & 31.39 & \textbf{42.25} & 37.35 & 37.93 & -6.53 \\
DISCO~\citep{guo2025federated} & 32.90 & 33.10 & 60.15 & 30.14 & 39.02 & 39.07 & 36.57 & \textbf{0.07}\\
DISCO$^*$~\citep{guo2025federated} & \textbf{34.20} & 35.00 & 61.55 & 27.50 & 40.14 & 39.56 & 37.85 & -0.77 \\
\rowcolor{Light}MR-LoRA~(Ours) & 33.70 & \textbf{36.20} & \textbf{65.10} & 32.50 & 41.89 & \textbf{41.88} & \textbf{38.86} & -0.02 \\
\bottomrule
\end{tabular}
}
\label{tab:llavaACL}
\vskip -0.1in
\end{table}

\begin{table}[t]
    \centering \small
    \setlength{\tabcolsep}{7.3pt}
    \caption{Results for InternVL-based domain continual learning in MLLM-CL benchmark. $^*$ denotes the original method with replay data.}  
    \vskip -0.1in
    \renewcommand{\arraystretch}{1.05}
\scalebox{0.85}{
\begin{tabular}{l|ccccc|cccc}
\toprule
Method     & RS      & Med      & AD     & Sci       & Fin      & MFT$\uparrow$ & MFN$\uparrow$ & MAA$\uparrow$ & BWT$\uparrow$ \\ 
\midrule
Zeroshot                            & 31.16 & 29.81 & 14.06 & 33.93 & 64.32 & 34.66 & -     & -     & -      \\
Oracle                              & 81.49 & 66.42 & 54.56 & 54.48 & 91.24 & {69.64} & -     & -     & -      \\
\midrule
\midrule
LoRA-FT~\citep{hu2021lora} & 69.93 & 52.17 & 33.04 & 42.67 & 91.07 & 69.06 & 57.78 & 65.22 & -14.11 \\
LoRA-FT$^*$~\citep{hu2021lora} & 77.06 & 47.55 & 42.67 & 43.31 & \textbf{91.44} & 69.43 & 60.41 & 67.45 & -11.28 \\
MoELoRA~\citep{chen2024coin} & 69.90 & 52.08 & 33.17 & 42.19 & 90.58 & 68.83 & 57.58 & 65.97 & -14.06 \\
MoELoRA$^*$~\citep{chen2024coin} & 76.74 & 52.65 & 38.81 & 42.15 & 89.84 & 67.90 & 60.04 & 66.01 & -9.83 \\
HiDe~\citep{guo2025hide} & 75.40 & 57.66 & 36.73 & 41.48 & 88.59 & 65.26 & 59.97 & 65.94 & -6.60 \\
HiDe$^*$~\citep{guo2025hide} & 53.17 & 52.61 & 40.85 & 47.04 & 89.17 & 64.20 & 56.57 & 61.06 & -9.54 \\
DISCO~\citep{guo2025federated} & 75.12 & 50.69 & 52.41 & 50.67 & 90.86 & 68.85 & 63.95 & 68.14 & -6.12 \\
DISCO$^*$~\citep{guo2025federated} & 77.90 & 47.50 & 49.13 & 49.37 & 90.92 & 68.55 & 62.96 & 67.81 & -6.98 \\
\rowcolor{Light}MR-LoRA~(Ours) & \textbf{81.48} & \textbf{65.80} & \textbf{54.56} & \textbf{54.40} & 91.07 & \textbf{69.51} & \textbf{69.46} & \textbf{71.27} & \textbf{-0.06} \\
\bottomrule
\end{tabular}
}
\label{tab:internvlDCL}
\end{table}
\begin{table}[t]
\centering \small
\setlength{\tabcolsep}{8pt}
\caption{Results for InternVL-based ability continual learning in MLLM-CL benchmark.}
\vskip -0.1in
\renewcommand{\arraystretch}{1.05}
\scalebox{0.85}{
\begin{tabular}{l|cccc|cccc}
\toprule
Method     & OCR   & M\&L  & VP    & GUI Agent & MFT$\uparrow$ & MFN$\uparrow$ & MAA$\uparrow$ & BWT$\uparrow$ \\
\midrule
Zeroshot                                & 30.00 & 31.20 & 56.09 & 2.50  & 29.95 & -     & -     & -     \\
Oracle                                  & 32.20 & 33.40 & 67.77 & 33.75 & 41.78 & -     & -     & -     \\
\midrule
\midrule
LoRA-FT~\citep{hu2021lora} & 21.40 & 32.80 & 60.28 & 29.86 & 40.84 & 36.08 & 36.38 & -6.35 \\
LoRA-FT$^*$~\citep{hu2021lora} & 26.30 & 34.20 & 62.56 & 31.25 & 41.63 & 38.58 & 37.38 & -4.07 \\
O-LoRA~\citep{wang2023orthogonal} & 25.50 & 32.30 & 64.59 & 24.44 & 38.64 & 36.71 & 36.05 & -2.57 \\
O-LoRA$^*$~\citep{wang2023orthogonal} & 21.70 & 31.10 & 59.77 & 31.25 & 41.38 & 35.96 & 36.49 & -7.23 \\
MoELoRA~\citep{chen2024coin} & 17.20 & 32.70 & 55.33 & 32.50 & 41.41 & 34.43 & 35.36 & -9.30 \\
MoELoRA$^*$~\citep{chen2024coin} & 13.90 & 29.70 & 54.95 & 32.50 & 41.91 & 32.76 & 35.66 & -12.20 \\
HiDe~\citep{guo2025hide} & 17.70 & 33.00 & 41.12 & 20.28 & 37.27 & 28.02 & 33.25 & -12.33 \\
HiDe$^*$~\citep{guo2025hide} & 25.30 & 29.20 & 42.13 & 20.28 & 35.93 & 29.23 & 33.39 & -8.93 \\
DISCO~\citep{guo2025federated} & 30.60 & 33.10 & 65.36 & 27.50 & 39.21 & 39.14 & 36.73 & -0.10 \\
DISCO$^*$~\citep{guo2025federated} & 32.30 & 32.30 & 64.97 & 30.14 & 40.46 & 39.93 & 37.63 & -0.71 \\
\rowcolor{Light}MR-LoRA~(Ours) & \textbf{33.00} & \textbf{35.70} & \textbf{67.51} & \textbf{33.75} & \textbf{42.56} & \textbf{42.49} & \textbf{38.85} & \textbf{-0.09} \\
\bottomrule
\end{tabular}
}
\label{tab:internvlACL}
\vskip -0.1in
\end{table}

\subsection{Results and Analysis}

\textbf{Domain Continual Learning.}
As demonstrated in \Cref{tab:llavaDCL} (LLaVA-based) and \Cref{tab:internvlDCL} (InternVL-based), our proposed MR-LoRA method achieves state-of-the-art performance on the DCL setting, showcasing its exceptional ability to acquire new domain knowledge while preserving previously learned capabilities.
The performance of MR-LoRA highlights several key advantages:
(1) \textbf{\textit{Approaching Oracle Performance:}} Our method's final accuracy on all individual tasks nearly matches the ``Oracle'' performance. For instance, in \Cref{tab:llavaDCL}, the final accuracies of MR-LoRA across the five domains are almost identical to the Oracle scores. This indicates that our MLLM-based router can select the most appropriate expert module for each input sample with high precision, allowing the overall performance to approach the theoretical upper bound of a perfect selection mechanism.
(2) \textbf{\textit{Superiority over Existing Baselines:}} In contrast, other baseline methods exhibit significant performance degradation.
Parameter-sharing and regularization methods like LoRA-FT and O-LoRA suffer from severe forgetting, as evidenced by their deeply negative BWT scores (\textit{e.g.}, -14.97 for LoRA-FT on LLaVA). This empirically confirms our hypothesis in \Cref{sec:conflict} regarding the severe task conflict among heterogeneous domains, where shared parameters compromise existing abilities while learning new ones.
Although replay-based methods (marked with~$^*$) alleviate forgetting by rehearsing old data, their performance remains far inferior to MR-LoRA. 
Even more advanced baselines like DISCO$^*$ and SEFE$^*$ still show a significant gap compared to ours.

\begin{figure}[t]
    \centering
    \includegraphics[width=\linewidth]{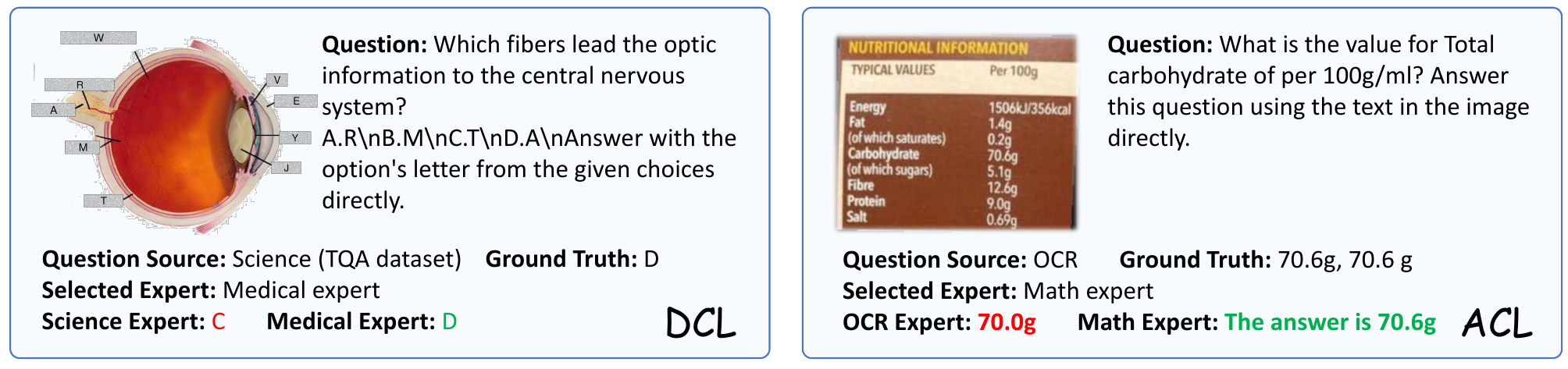}
    \vskip -0.1in
    \caption{Examples demonstrating that the selected expert handles certain questions better than the original expert in DCL and ACL. MLLM-enhanced router selects the most appropriate experts.}
    \label{fig:routerexam}
\end{figure}

\textbf{Ability Continual Learning.}
The effectiveness of our proposed method in the more challenging ACL setting is demonstrated in \Cref{tab:llavaACL,tab:internvlACL}. This setting evaluates the model's capacity to acquire fundamental new skills and generalize to non-IID test sets. Firstly, we observe that most baselines suffer from severe catastrophic forgetting, 
revealing a critical weakness in existing CL approaches when faced with real-world, practical non-IID scenarios. In contrast, our MR-LoRA significantly outperforms all baseline methods and successfully improves performance across all four abilities by isolating abilities into dedicated expert modules and leveraging an intelligent MLLM-based router.
\begin{wraptable}{r}{6.7cm}
\vskip -0in
  \centering \small
    \setlength{\tabcolsep}{4pt}
    \caption{Ablation study of LoRA rank for each expert LoRA (LLaVA, DCL, last accuracy).}
    \vskip -0.1in
    \renewcommand{\arraystretch}{1.1}
    \scalebox{0.9}{
    \begin{tabular}{r|ccccc}
    \toprule
        Rank & RS (\%) & Med (\%) & AD (\%) &Sci (\%)&Fin (\%) \\
        \midrule
8   & 80.96 & 64.64 & 54.00    & 55.44 & 90.75 \\
16  & 80.92 & 65.11 & 53.98 & 55.90  & 91.02 \\
  \rowcolor{Light} 32  & 80.87 & 65.32 & 54.12 & 56.71 & 91.12 \\
64  & 81.18 & 66.07 & 54.31 & 56.90  & 91.60  \\
128 & 81.14 & 66.49 & 54.00    & 57.63 & 91.44 \\
         \bottomrule
    \end{tabular}}\label{tab:rank}
    \vskip -0in
\end{wraptable}
Interestingly, the results also reveal a knowledge transfer enabled by our MLLM-enhanced router. In the InternVL experiments, the final accuracy of MR-LoRA on the OCR task is 33.00\%, which is higher than the 32.20\% achieved by the Oracle.
This suggests that the router's flexible selection mechanism can sometimes leverage knowledge from other related experts (\textit{e.g.}, using the OCR capabilities in the M \& L expert) to achieve a result superior to that of a single, isolated specialist. 
This phenomenon highlights the rationality and sophisticated decision-making capability of the MR-LoRA framework.
\Cref{fig:routerexam} shows the knowledge transfer phenomenon in DCL and ACL.

\textbf{Rank of Expert LoRA.}
From the results in \Cref{tab:rank}, we find that our method performs well even at very low ranks (\textit{e.g.}, 8), demonstrating its parameter efficiency.
This indicates that even if the number of tasks to be learned is large, our method can still achieve a good performance with only a small increase in parameters.
Besides, as the expert rank increases, performance can be improved slightly because of more trainable parameters.

\textbf{Router Accuracy.}
We ablate the number of samples for routing data and report the router selection accuracy and the last accuracy in domain and ability continual learning.
The results are shown in \Cref{tab:routerdcl,tab:routeracl}.
In DCL, we find that our method can achieve an excellent performance (almost 100\% selection accuracy) using only 20 samples to train the router, which means our method closes the gap of training each task individually.
Note that the number of samples we used is much smaller than the number of training samples (60k).
Besides, with more sampling data, the router selection accuracy improves and the performance of~\ours~slightly increases.
In ACL, the performance of~\ours~achieves satisfactory performance when the shot of router tuning is 10.
It is interesting that the router accuracy of the OCR task is around 50\%, but our method can achieve a comparable, or even better performance compared with directly finetuning an OCR LoRA expert (33.60\%).
This means~\ours~uses other experts to solve the OCR task, and these experts perform well on these test samples.
It is reasonable that OCR is a basic and fundamental ability that the math and GUI Agent experts are also able to extract equations and web texts from the images.

\begin{table}[t]
\centering \small
\captionof{table}{Router accuracy under different amount of router data in \textit{domain continual learning}.
The left part is the router selection accuracy and the right part is task accuracy after learning the last task.}
\vskip -0.1in
\renewcommand{\arraystretch}{1.03}
\scalebox{0.9}{
\begin{tabular}{r|cccccccccc}
\toprule
\multirow{2}{*}{\begin{tabular}[c]{@{}c@{}}\# Replay\\ Samples\end{tabular}} & \multicolumn{5}{c}{Router Accuracy (\%)}  & \multicolumn{5}{c}{Last Accuracy (\%)} \\
\cmidrule(lr){2-6} \cmidrule(lr){7-11}
& RS & Med & AD & Sci & Fin & RS & Med & AD & Sci & Fin\\
\midrule
100 & 99.96 & 99.16 & 99.98  & 98.44 & 99.99  & 81.04 & 65.61 & 54.16 & 56.77 & 91.13 \\
50  & 99.85 & 98.69 & 99.94  & 98.82 & 100.00 & 81.00 & 65.53 & 54.14 & 56.76 & 91.14 \\
30  & 99.62 & 98.89 & 100.00 & 96.90 & 99.86  & 80.92 & 65.53 & 54.17 & 56.59 & 91.08 \\
\rowcolor{Light}
20  & 99.52 & 97.87 & 99.89  & 98.40 & 99.80  & 80.87 & 65.32 & 54.12 & 56.71 & 91.12 \\
10  & 99.93 & 98.24 & 99.93  & 97.75 & 99.40  & 81.04 & 65.40 & 54.16 & 56.63 & 91.01 \\
\bottomrule
\end{tabular}
}
\label{tab:routerdcl}
\vskip -0.1in
\end{table}

\begin{table}[!t]
\centering \small
\setlength{\tabcolsep}{7.5pt}
\caption{\centering Router accuracy under different amount of replay data in \textit{ability continual learning}.}
\vskip -0.05in
\renewcommand{\arraystretch}{1.03}
\scalebox{0.9}{
\begin{tabular}{r|cccccccc}
\toprule
\multirow{2}{*}{\begin{tabular}[c]{@{}c@{}}\# Replay\\ Samples\end{tabular}} & \multicolumn{4}{c}{Router Accuracy (\%)}  & \multicolumn{4}{c}{Last Accuracy (\%)} \\
\cmidrule(lr){2-5} \cmidrule(lr){6-9}
& OCR & M\&L & VP & GUI Agent & OCR & M\&L & VP & GUI Agent \\
\midrule
100 & 72.10 & 94.60 & 99.87  & 100.00 & 32.80 & 36.30 & 65.10 & 32.50 \\
50  & 65.30 & 83.90 & 99.11  & 100.00 & 32.70 & 36.10 & 64.85 & 32.50 \\
30  & 53.60 & 90.90 & 97.21  & 98.38  & 33.80 & 36.70 & 64.85 & 32.50 \\
\rowcolor{Light}
20  & 51.40 & 86.00 & 100.00 & 100.00 & 33.70 & 36.20 & 65.10 & 32.50 \\
10  & 81.90 & 76.30 & 100.00 & 100.00 & 32.80 & 35.80 & 65.10 & 32.50 \\
\bottomrule
\end{tabular}
}
\label{tab:routeracl}
\vskip -0.1in
\end{table}

   

\section{Conclusion}
In this paper, we first propose MLLM-CL benchmark, a novel benchmark including domain continual learning and ability continual learning.
In domain continual learning, we select five specific domains (remote sensing, medical, science, autonomous driving, and finance) and focus on IID evaluation.
In ability continual learning, we consider a more practical setting where the training and test sets are non-IID.
We select four common and fundamental abilities for MLLM to learning sequentially: OCR, math \& logic, visual perception, and GUI agent.
To solve the two settings in the MLLM-CL benchmark, we first analyze the task conflict between different tasks and then propose an MLLM enhanced router selection method~\ours.
Comprehensive experiments and analyses validate the necessity of our MLLM-CL benchmark and show the effectiveness and efficiency of our proposed method.
We believe that our carefully designed benchmark and~\ours~can serve as a foundation for continual learning in multimodal large language models and will introduce an innovative and practical direction of continual learning and MLLM to the community.

\newpage

\section*{Ethics statement}
Our research is grounded in ethical practices, with particular attention paid to the responsible use of data. 
This work exclusively employs public, well-established datasets from the MLLM community, and we list all used assets' licenses in \Cref{tab:license_reorganized}.
Our use of this data is in accordance with their provided licenses and intended academic purpose.

\section*{Reproducibility statement}
To facilitate the reproducibility of our research, we provide comprehensive implementation details in \Cref{sec:implement}, including training procedures and hyperparameters. We also report all the result matrices in \Cref{sec:bigmatrix}. 
All source code, datasets, and trained models will be publicly released upon the paper's acceptance.

\bibliography{iclr2026_conference}

\begin{thebibliography}{69}
\providecommand{\natexlab}[1]{#1}
\providecommand{\url}[1]{\texttt{#1}}
\expandafter\ifx\csname urlstyle\endcsname\relax
  \providecommand{\doi}[1]{doi: #1}\else
  \providecommand{\doi}{doi: \begingroup \urlstyle{rm}\Url}\fi

\bibitem[Acharya et~al.(2019)Acharya, Kafle, and Kanan]{acharya2019tallyqa}
Manoj Acharya, Kushal Kafle, and Christopher Kanan.
\newblock Tallyqa: Answering complex counting questions.
\newblock In \emph{AAAI}, 2019.

\bibitem[Achiam et~al.(2023)Achiam, Adler, Agarwal, Ahmad, Akkaya, Aleman, Almeida, Altenschmidt, Altman, Anadkat, et~al.]{achiam2023gpt}
Josh Achiam, Steven Adler, Sandhini Agarwal, Lama Ahmad, Ilge Akkaya, Florencia~Leoni Aleman, Diogo Almeida, Janko Altenschmidt, Sam Altman, Shyamal Anadkat, et~al.
\newblock Gpt-4 technical report.
\newblock \emph{arXiv preprint arXiv:2303.08774}, 2023.

\bibitem[Bai et~al.(2023)Bai, Bai, Yang, Wang, Tan, Wang, Lin, Zhou, and Zhou]{bai2023qwen}
Jinze Bai, Shuai Bai, Shusheng Yang, Shijie Wang, Sinan Tan, Peng Wang, Junyang Lin, Chang Zhou, and Jingren Zhou.
\newblock Qwen-vl: A frontier large vision-language model with versatile abilities.
\newblock \emph{arXiv preprint arXiv:2308.12966}, 2023.

\bibitem[Bai et~al.(2025)Bai, Chen, Liu, Wang, Ge, Song, Dang, Wang, Wang, Tang, Zhong, Zhu, Yang, Li, Wan, Wang, Ding, Fu, Xu, Ye, Zhang, Xie, Cheng, Zhang, Yang, Xu, and Lin]{Qwen2.5-VL}
Shuai Bai, Keqin Chen, Xuejing Liu, Jialin Wang, Wenbin Ge, Sibo Song, Kai Dang, Peng Wang, Shijie Wang, Jun Tang, Humen Zhong, Yuanzhi Zhu, Mingkun Yang, Zhaohai Li, Jianqiang Wan, Pengfei Wang, Wei Ding, Zheren Fu, Yiheng Xu, Jiabo Ye, Xi~Zhang, Tianbao Xie, Zesen Cheng, Hang Zhang, Zhibo Yang, Haiyang Xu, and Junyang Lin.
\newblock Qwen2.5-vl technical report.
\newblock \emph{arXiv preprint arXiv:2502.13923}, 2025.

\bibitem[Buzzega et~al.(2020)Buzzega, Boschini, Porrello, Abati, and Calderara]{buzzega2020dark}
Pietro Buzzega, Matteo Boschini, Angelo Porrello, Davide Abati, and Simone Calderara.
\newblock Dark experience for general continual learning: a strong, simple baseline.
\newblock \emph{Advances in neural information processing systems}, 33:\penalty0 15920--15930, 2020.

\bibitem[Cao et~al.(2024)Cao, Liu, Liu, Wang, Dong, Ding, Zhang, Reid, and Liang]{cao2024continual}
Meng Cao, Yuyang Liu, Yingfei Liu, Tiancai Wang, Jiahua Dong, Henghui Ding, Xiangyu Zhang, Ian Reid, and Xiaodan Liang.
\newblock Continual llava: Continual instruction tuning in large vision-language models.
\newblock \emph{arXiv preprint arXiv:2411.02564}, 2024.

\bibitem[Chang et~al.(2022)Chang, Palzer, Li, Fosler-Lussier, and Xiao]{chang2022mapqa}
Shuaichen Chang, David Palzer, Jialin Li, Eric Fosler-Lussier, and Ningchuan Xiao.
\newblock Mapqa: A dataset for question answering on choropleth maps.
\newblock \emph{arXiv preprint arXiv:2211.08545}, 2022.

\bibitem[Chen et~al.(2024{\natexlab{a}})Chen, Zhu, Luo, Shen, Song, and Gao]{chen2024coin}
Cheng Chen, Junchen Zhu, Xu~Luo, Hengtao Shen, Jingkuan Song, and Lianli Gao.
\newblock Coin: A benchmark of continual instruction tuning for multimodel large language models.
\newblock \emph{Advances in Neural Information Processing Systems}, 37:\penalty0 57817--57840, 2024{\natexlab{a}}.

\bibitem[Chen et~al.(2024{\natexlab{b}})Chen, Chen, Zhang, Chen, Wu, Zhang, Chen, Li, Wan, and Wang]{chen2024allava}
Guiming~Hardy Chen, Shunian Chen, Ruifei Zhang, Junying Chen, Xiangbo Wu, Zhiyi Zhang, Zhihong Chen, Jianquan Li, Xiang Wan, and Benyou Wang.
\newblock Allava: Harnessing gpt4v-synthesized data for lite vision-language models.
\newblock \emph{arXiv preprint arXiv:2402.11684}, 2024{\natexlab{b}}.

\bibitem[Chen et~al.(2025)Chen, Cong, Zhao, Yang, Hu, Ip, and Kwong]{chen2025sefe}
Jinpeng Chen, Runmin Cong, Yuzhi Zhao, Hongzheng Yang, Guangneng Hu, Horace Ip, and Sam Kwong.
\newblock Sefe: Superficial and essential forgetting eliminator for multimodal continual instruction tuning.
\newblock In \emph{Forty-second International Conference on Machine Learning}, 2025.

\bibitem[Chen et~al.(2024{\natexlab{c}})Chen, Wang, Cao, Liu, Gao, Cui, Zhu, Ye, Tian, Liu, et~al.]{chen2024expanding}
Zhe Chen, Weiyun Wang, Yue Cao, Yangzhou Liu, Zhangwei Gao, Erfei Cui, Jinguo Zhu, Shenglong Ye, Hao Tian, Zhaoyang Liu, et~al.
\newblock Expanding performance boundaries of open-source multimodal models with model, data, and test-time scaling.
\newblock \emph{arXiv preprint arXiv:2412.05271}, 2024{\natexlab{c}}.

\bibitem[Chen et~al.(2024{\natexlab{d}})Chen, Wu, Wang, Su, Chen, Xing, Zhong, Zhang, Zhu, Lu, et~al.]{chen2024internvl}
Zhe Chen, Jiannan Wu, Wenhai Wang, Weijie Su, Guo Chen, Sen Xing, Muyan Zhong, Qinglong Zhang, Xizhou Zhu, Lewei Lu, et~al.
\newblock Internvl: Scaling up vision foundation models and aligning for generic visual-linguistic tasks.
\newblock In \emph{Proceedings of the IEEE/CVF Conference on Computer Vision and Pattern Recognition}, pp.\  24185--24198, 2024{\natexlab{d}}.

\bibitem[Cossu et~al.(2024)Cossu, Carta, Passaro, Lomonaco, Tuytelaars, and Bacciu]{cossu2024continual}
Andrea Cossu, Antonio Carta, Lucia Passaro, Vincenzo Lomonaco, Tinne Tuytelaars, and Davide Bacciu.
\newblock Continual pre-training mitigates forgetting in language and vision.
\newblock \emph{Neural Networks}, 179:\penalty0 106492, 2024.

\bibitem[Douillard et~al.(2022)Douillard, Ram{\'e}, Couairon, and Cord]{Douillard_Rame_Couairon_Cord_2022}
Arthur Douillard, Alexandre Ram{\'e}, Guillaume Couairon, and Matthieu Cord.
\newblock Dytox: Transformers for continual learning with dynamic token expansion.
\newblock In \emph{Proceedings of the IEEE/CVF Conference on Computer Vision and Pattern Recognition}, pp.\  9285--9295, 2022.

\bibitem[Fu et~al.(2025)Fu, Lin, Wang, Zhang, Shen, Liu, Li, Long, Gao, Li, et~al.]{fu2025vita}
Chaoyou Fu, Haojia Lin, Xiong Wang, Yi-Fan Zhang, Yunhang Shen, Xiaoyu Liu, Yangze Li, Zuwei Long, Heting Gao, Ke~Li, et~al.
\newblock Vita-1.5: Towards gpt-4o level real-time vision and speech interaction.
\newblock \emph{arXiv preprint arXiv:2501.01957}, 2025.

\bibitem[Goyal et~al.(2017)Goyal, Khot, Summers-Stay, Batra, and Parikh]{goyal2017making}
Yash Goyal, Tejas Khot, Douglas Summers-Stay, Dhruv Batra, and Devi Parikh.
\newblock Making the v in vqa matter: Elevating the role of image understanding in visual question answering.
\newblock In \emph{Proceedings of the IEEE conference on computer vision and pattern recognition}, pp.\  6904--6913, 2017.

\bibitem[Guo et~al.(2025{\natexlab{a}})Guo, Zeng, Xiang, Zhu, Wang, Zhang, and Liu]{guo2025hide}
Haiyang Guo, Fanhu Zeng, Ziwei Xiang, Fei Zhu, Da-Han Wang, Xu-Yao Zhang, and Cheng-Lin Liu.
\newblock Hide-llava: Hierarchical decoupling for continual instruction tuning of multimodal large language model.
\newblock \emph{arXiv preprint arXiv:2503.12941}, 2025{\natexlab{a}}.

\bibitem[Guo et~al.(2025{\natexlab{b}})Guo, Zeng, Zhu, Liu, Wang, Xu, Zhang, and Liu]{guo2025federated}
Haiyang Guo, Fanhu Zeng, Fei Zhu, Wenzhuo Liu, Da-Han Wang, Jian Xu, Xu-Yao Zhang, and Cheng-Lin Liu.
\newblock Federated continual instruction tuning.
\newblock \emph{arXiv preprint arXiv:2503.12897}, 2025{\natexlab{b}}.

\bibitem[Guo et~al.(2025{\natexlab{c}})Guo, Zeng, Zhu, Wang, Wang, Zhou, Zhao, Liu, Ma, Wang, et~al.]{guo2025comprehensive}
Haiyang Guo, Fanhu Zeng, Fei Zhu, Jiayi Wang, Xukai Wang, Jingang Zhou, Hongbo Zhao, Wenzhuo Liu, Shijie Ma, Da-Han Wang, et~al.
\newblock A comprehensive survey on continual learning in generative models.
\newblock \emph{arXiv preprint arXiv:2506.13045}, 2025{\natexlab{c}}.

\bibitem[Guo et~al.(2025{\natexlab{d}})Guo, Zhu, Zhao, Zeng, Liu, Ma, Wang, and Zhang]{guo2025mcitlib}
Haiyang Guo, Fei Zhu, Hongbo Zhao, Fanhu Zeng, Wenzhuo Liu, Shijie Ma, Da-Han Wang, and Xu-Yao Zhang.
\newblock Mcitlib: Multimodal continual instruction tuning library and benchmark.
\newblock \emph{arXiv preprint arXiv:2508.07307}, 2025{\natexlab{d}}.

\bibitem[Guo et~al.(2025{\natexlab{e}})Guo, Zhang, Chen, Gao, Jiang, Wang, and Heng]{guo2025sciverse}
Ziyu Guo, Ray Zhang, Hao Chen, Jialin Gao, Dongzhi Jiang, Jiaze Wang, and Pheng-Ann Heng.
\newblock Sciverse: Unveiling the knowledge comprehension and visual reasoning of lmms on multi-modal scientific problems.
\newblock \emph{arXiv preprint arXiv:2503.10627}, 2025{\natexlab{e}}.

\bibitem[He et~al.(2023)He, Guo, Tang, and Wang]{he2023continual}
Jinghan He, Haiyun Guo, Ming Tang, and Jinqiao Wang.
\newblock Continual instruction tuning for large multimodal models.
\newblock \emph{arXiv preprint arXiv:2311.16206}, 2023.

\bibitem[He et~al.(2020)He, Zhang, Mou, Xing, and Xie]{he2020pathvqa}
Xuehai He, Yichen Zhang, Luntian Mou, Eric Xing, and Pengtao Xie.
\newblock Pathvqa: 30000+ questions for medical visual question answering.
\newblock \emph{arXiv preprint arXiv:2003.10286}, 2020.

\bibitem[Hsiao et~al.(2022)Hsiao, Zubach, Baechler, Carbune, Lin, Wang, Sunkara, Zhu, and Chen]{hsiao2022screenqa}
Yu-Chung Hsiao, Fedir Zubach, Gilles Baechler, Victor Carbune, Jason Lin, Maria Wang, Srinivas Sunkara, Yun Zhu, and Jindong Chen.
\newblock Screenqa: Large-scale question-answer pairs over mobile app screenshots.
\newblock \emph{arXiv preprint arXiv:2209.08199}, 2022.

\bibitem[Hu et~al.(2021)Hu, Shen, Wallis, Allen-Zhu, Li, Wang, Wang, and Chen]{hu2021lora}
Edward~J Hu, Yelong Shen, Phillip Wallis, Zeyuan Allen-Zhu, Yuanzhi Li, Shean Wang, Lu~Wang, and Weizhu Chen.
\newblock Lora: Low-rank adaptation of large language models.
\newblock \emph{arXiv preprint arXiv:2106.09685}, 2021.

\bibitem[Huai et~al.(2025)Huai, Zhou, Wu, Chen, Bai, Zhou, and He]{huai2025cl}
Tianyu Huai, Jie Zhou, Xingjiao Wu, Qin Chen, Qingchun Bai, Ze~Zhou, and Liang He.
\newblock Cl-moe: Enhancing multimodal large language model with dual momentum mixture-of-experts for continual visual question answering.
\newblock \emph{arXiv preprint arXiv:2503.00413}, 2025.

\bibitem[Huang et~al.(2024)Huang, Cao, Lu, and Liu]{huang2024class}
Linlan Huang, Xusheng Cao, Haori Lu, and Xialei Liu.
\newblock Class-incremental learning with clip: Adaptive representation adjustment and parameter fusion.
\newblock In \emph{European Conference on Computer Vision}, pp.\  214--231. Springer, 2024.

\bibitem[Jang et~al.(2022)Jang, Ye, Lee, Yang, Shin, Han, Kim, and Seo]{jang2022temporalwiki}
Joel Jang, Seonghyeon Ye, Changho Lee, Sohee Yang, Joongbo Shin, Janghoon Han, Gyeonghun Kim, and Minjoon Seo.
\newblock Temporalwiki: A lifelong benchmark for training and evaluating ever-evolving language models.
\newblock \emph{arXiv preprint arXiv:2204.14211}, 2022.

\bibitem[Johnson et~al.(2017)Johnson, Hariharan, Van Der~Maaten, Fei-Fei, Lawrence~Zitnick, and Girshick]{johnson2017clevr}
Justin Johnson, Bharath Hariharan, Laurens Van Der~Maaten, Li~Fei-Fei, C~Lawrence~Zitnick, and Ross Girshick.
\newblock Clevr: A diagnostic dataset for compositional language and elementary visual reasoning.
\newblock In \emph{Proceedings of the IEEE conference on computer vision and pattern recognition}, pp.\  2901--2910, 2017.

\bibitem[Kembhavi et~al.(2016)Kembhavi, Salvato, Kolve, Seo, Hajishirzi, and Farhadi]{kembhavi2016diagram}
Aniruddha Kembhavi, Mike Salvato, Eric Kolve, Minjoon Seo, Hannaneh Hajishirzi, and Ali Farhadi.
\newblock A diagram is worth a dozen images.
\newblock In \emph{Computer Vision--ECCV 2016: 14th European Conference, Amsterdam, The Netherlands, October 11--14, 2016, Proceedings, Part IV 14}, pp.\  235--251. Springer, 2016.

\bibitem[Kembhavi et~al.(2017)Kembhavi, Seo, Schwenk, Choi, Farhadi, and Hajishirzi]{kembhavi2017you}
Aniruddha Kembhavi, Minjoon Seo, Dustin Schwenk, Jonghyun Choi, Ali Farhadi, and Hannaneh Hajishirzi.
\newblock Are you smarter than a sixth grader? textbook question answering for multimodal machine comprehension.
\newblock In \emph{Proceedings of the IEEE Conference on Computer Vision and Pattern recognition}, pp.\  4999--5007, 2017.

\bibitem[Kirkpatrick et~al.(2017)Kirkpatrick, Pascanu, Rabinowitz, Veness, Desjardins, Rusu, Milan, Quan, Ramalho, Grabska-Barwinska, et~al.]{kirkpatrick2017overcoming}
James Kirkpatrick, Razvan Pascanu, Neil Rabinowitz, Joel Veness, Guillaume Desjardins, Andrei~A Rusu, Kieran Milan, John Quan, Tiago Ramalho, Agnieszka Grabska-Barwinska, et~al.
\newblock Overcoming catastrophic forgetting in neural networks.
\newblock \emph{Proceedings of the national academy of sciences}, 114\penalty0 (13):\penalty0 3521--3526, 2017.

\bibitem[Kwon et~al.(2023)Kwon, Li, Zhuang, Sheng, Zheng, Yu, Gonzalez, Zhang, and Stoica]{kwon2023efficient}
Woosuk Kwon, Zhuohan Li, Siyuan Zhuang, Ying Sheng, Lianmin Zheng, Cody~Hao Yu, Joseph~E. Gonzalez, Hao Zhang, and Ion Stoica.
\newblock Efficient memory management for large language model serving with pagedattention.
\newblock In \emph{Proceedings of the ACM SIGOPS 29th Symposium on Operating Systems Principles}, 2023.

\bibitem[Lavda et~al.(2018)Lavda, Ramapuram, Gregorova, and Kalousis]{lavda2018continual}
Frantzeska Lavda, Jason Ramapuram, Magda Gregorova, and Alexandros Kalousis.
\newblock Continual classification learning using generative models.
\newblock \emph{arXiv preprint arXiv:1810.10612}, 2018.

\bibitem[Li et~al.(2024{\natexlab{a}})Li, Zhang, Guo, Zhang, Li, Zhang, Zhang, Li, Liu, and Li]{li2024llava}
Bo~Li, Yuanhan Zhang, Dong Guo, Renrui Zhang, Feng Li, Hao Zhang, Kaichen Zhang, Yanwei Li, Ziwei Liu, and Chunyuan Li.
\newblock Llava-onevision: Easy visual task transfer.
\newblock \emph{arXiv preprint arXiv:2408.03326}, 2024{\natexlab{a}}.

\bibitem[Li et~al.(2024{\natexlab{b}})Li, Yang, Liu, Ma, Zhang, Yang, Sun, Liu, and Bai]{li2023monkey}
Zhang Li, Biao Yang, Qiang Liu, Zhiyin Ma, Shuo Zhang, Jingxu Yang, Yabo Sun, Yuliang Liu, and Xiang Bai.
\newblock Monkey: Image resolution and text label are important things for large multi-modal models.
\newblock In \emph{proceedings of the IEEE/CVF conference on computer vision and pattern recognition}, 2024{\natexlab{b}}.

\bibitem[Li \& Hoiem(2017)Li and Hoiem]{li2017learning}
Zhizhong Li and Derek Hoiem.
\newblock Learning without forgetting.
\newblock \emph{IEEE transactions on pattern analysis and machine intelligence}, 40\penalty0 (12):\penalty0 2935--2947, 2017.

\bibitem[Liu et~al.(2023)Liu, Li, Wu, and Lee]{liu2023llava}
Haotian Liu, Chunyuan Li, Qingyang Wu, and Yong~Jae Lee.
\newblock Visual instruction tuning.
\newblock \emph{Advances in neural information processing systems}, 36:\penalty0 34892--34916, 2023.

\bibitem[Liu et~al.(2024{\natexlab{a}})Liu, Li, Li, and Lee]{liu2023improvedllava}
Haotian Liu, Chunyuan Li, Yuheng Li, and Yong~Jae Lee.
\newblock Improved baselines with visual instruction tuning.
\newblock In \emph{Proceedings of the IEEE/CVF Conference on Computer Vision and Pattern Recognition}, pp.\  26296--26306, 2024{\natexlab{a}}.

\bibitem[Liu et~al.(2024{\natexlab{b}})Liu, Ou, Song, Qu, Lam, Xiong, Chen, Neubig, and Yue]{liu2024harnessing}
Junpeng Liu, Tianyue Ou, Yifan Song, Yuxiao Qu, Wai Lam, Chenyan Xiong, Wenhu Chen, Graham Neubig, and Xiang Yue.
\newblock Harnessing webpage uis for text-rich visual understanding.
\newblock \emph{arXiv preprint arXiv:2410.13824}, 2024{\natexlab{b}}.

\bibitem[Liu et~al.(2024{\natexlab{c}})Liu, Lin, Hewitt, Paranjape, Bevilacqua, Petroni, and Liang]{liu2024lost}
Nelson~F Liu, Kevin Lin, John Hewitt, Ashwin Paranjape, Michele Bevilacqua, Fabio Petroni, and Percy Liang.
\newblock Lost in the middle: How language models use long contexts.
\newblock \emph{Transactions of the Association for Computational Linguistics}, 12:\penalty0 157--173, 2024{\natexlab{c}}.

\bibitem[Liu et~al.(2024{\natexlab{d}})Liu, Li, Huang, Yang, Yu, Li, Yin, Liu, Jin, and Bai]{OCRBench}
Yuliang Liu, Zhang Li, Mingxin Huang, Biao Yang, Wenwen Yu, Chunyuan Li, Xu-Cheng Yin, Cheng-Lin Liu, Lianwen Jin, and Xiang Bai.
\newblock Ocrbench: on the hidden mystery of ocr in large multimodal models.
\newblock \emph{Science China Information Sciences}, 67\penalty0 (12), December 2024{\natexlab{d}}.
\newblock ISSN 1869-1919.
\newblock \doi{10.1007/s11432-024-4235-6}.
\newblock URL \url{http://dx.doi.org/10.1007/s11432-024-4235-6}.

\bibitem[Lobry et~al.(2020)Lobry, Marcos, Murray, and Tuia]{lobry2020rsvqa}
Sylvain Lobry, Diego Marcos, Jesse Murray, and Devis Tuia.
\newblock Rsvqa: Visual question answering for remote sensing data.
\newblock \emph{IEEE Transactions on Geoscience and Remote Sensing}, 58\penalty0 (12):\penalty0 8555--8566, 2020.

\bibitem[Lu et~al.(2024)Lu, Bansal, Xia, Liu, Li, Hajishirzi, Cheng, Chang, Galley, and Gao]{lu2024mathvista}
Pan Lu, Hritik Bansal, Tony Xia, Jiacheng Liu, Chunyuan Li, Hannaneh Hajishirzi, Hao Cheng, Kai-Wei Chang, Michel Galley, and Jianfeng Gao.
\newblock Mathvista: Evaluating mathematical reasoning of foundation models in visual contexts.
\newblock In \emph{International Conference on Learning Representations (ICLR)}, 2024.

\bibitem[Mallya et~al.(2018)Mallya, Davis, and Lazebnik]{mallya2018piggyback}
Arun Mallya, Dillon Davis, and Svetlana Lazebnik.
\newblock Piggyback: Adapting a single network to multiple tasks by learning to mask weights.
\newblock In \emph{Proceedings of the European conference on computer vision (ECCV)}, pp.\  67--82, 2018.

\bibitem[McCloskey \& Cohen(1989)McCloskey and Cohen]{mccloskey1989catastrophic}
Michael McCloskey and Neal~J Cohen.
\newblock Catastrophic interference in connectionist networks: The sequential learning problem.
\newblock In \emph{Psychology of learning and motivation}, volume~24, pp.\  109--165. Elsevier, 1989.

\bibitem[OpenAI(2024)]{gpt4o}
OpenAI.
\newblock Hello gpt-4o.
\newblock \url{https://openai.com/index/hello-gpt-4o/}, 2024.

\bibitem[Radford et~al.(2021)Radford, Kim, Hallacy, Ramesh, Goh, Agarwal, Sastry, Askell, Mishkin, Clark, et~al.]{radford2021learning}
Alec Radford, Jong~Wook Kim, Chris Hallacy, Aditya Ramesh, Gabriel Goh, Sandhini Agarwal, Girish Sastry, Amanda Askell, Pamela Mishkin, Jack Clark, et~al.
\newblock Learning transferable visual models from natural language supervision.
\newblock In \emph{International conference on machine learning}, pp.\  8748--8763. PMLR, 2021.

\bibitem[Razdaibiedina et~al.(2023)Razdaibiedina, Mao, Hou, Khabsa, Lewis, and Almahairi]{razdaibiedina2023progressive}
Anastasia Razdaibiedina, Yuning Mao, Rui Hou, Madian Khabsa, Mike Lewis, and Amjad Almahairi.
\newblock Progressive prompts: Continual learning for language models.
\newblock \emph{arXiv preprint arXiv:2301.12314}, 2023.

\bibitem[Shi et~al.(2024{\natexlab{a}})Shi, Xu, Wang, Qin, Wang, Wang, Wang, Ebrahimi, and Wang]{shi2024continual}
Haizhou Shi, Zihao Xu, Hengyi Wang, Weiyi Qin, Wenyuan Wang, Yibin Wang, Zifeng Wang, Sayna Ebrahimi, and Hao Wang.
\newblock Continual learning of large language models: A comprehensive survey.
\newblock \emph{arXiv preprint arXiv:2404.16789}, 2024{\natexlab{a}}.

\bibitem[Shi et~al.(2024{\natexlab{b}})Shi, Hu, Bin, Liu, Yang, Ng, Bing, and Lee]{shihu2024mathllava}
Wenhao Shi, Zhiqiang Hu, Yi~Bin, Junhua Liu, Yang Yang, See-Kiong Ng, Lidong Bing, and Roy Ka-Wei Lee.
\newblock Math-llava: Bootstrapping mathematical reasoning for multimodal large language models.
\newblock \emph{arXiv preprint arXiv:2406.17294}, 2024{\natexlab{b}}.

\bibitem[Sima et~al.(2023)Sima, Renz, Chitta, Chen, Zhang, Xie, Luo, Geiger, and Li]{sima2023drivelm}
Chonghao Sima, Katrin Renz, Kashyap Chitta, Li~Chen, Hanxue Zhang, Chengen Xie, Ping Luo, Andreas Geiger, and Hongyang Li.
\newblock Drivelm: Driving with graph visual question answering.
\newblock \emph{arXiv preprint arXiv:2312.14150}, 2023.

\bibitem[Smith et~al.(2023)Smith, Karlinsky, Gutta, Cascante-Bonilla, Kim, Arbelle, Panda, Feris, and Kira]{smith2023coda}
James~Seale Smith, Leonid Karlinsky, Vyshnavi Gutta, Paola Cascante-Bonilla, Donghyun Kim, Assaf Arbelle, Rameswar Panda, Rogerio Feris, and Zsolt Kira.
\newblock Coda-prompt: Continual decomposed attention-based prompting for rehearsal-free continual learning.
\newblock In \emph{Proceedings of the IEEE/CVF Conference on Computer Vision and Pattern Recognition}, pp.\  11909--11919, 2023.

\bibitem[Suhr \& Artzi(2024)Suhr and Artzi]{suhr2024continual}
Alane Suhr and Yoav Artzi.
\newblock Continual learning for instruction following from realtime feedback.
\newblock \emph{Advances in Neural Information Processing Systems}, 36, 2024.

\bibitem[Tong et~al.(2024)Tong, Brown, Wu, Woo, Middepogu, Akula, Yang, Yang, Iyer, Pan, et~al.]{tong2024cambrian}
Shengbang Tong, Ellis Brown, Penghao Wu, Sanghyun Woo, Manoj Middepogu, Sai~Charitha Akula, Jihan Yang, Shusheng Yang, Adithya Iyer, Xichen Pan, et~al.
\newblock Cambrian-1: A fully open, vision-centric exploration of multimodal llms.
\newblock \emph{arXiv preprint arXiv:2406.16860}, 2024.

\bibitem[Wang et~al.(2021)Wang, Li, Zhou, Chen, Grossman, and Li]{wang2021screen2words}
Bryan Wang, Gang Li, Xin Zhou, Zhourong Chen, Tovi Grossman, and Yang Li.
\newblock Screen2words: Automatic mobile ui summarization with multimodal learning.
\newblock In \emph{The 34th Annual ACM Symposium on User Interface Software and Technology}, pp.\  498--510, 2021.

\bibitem[Wang et~al.(2023{\natexlab{a}})Wang, Chen, Ge, Xia, Bao, Zheng, Zhang, Gui, and Huang]{wang2023orthogonal}
Xiao Wang, Tianze Chen, Qiming Ge, Han Xia, Rong Bao, Rui Zheng, Qi~Zhang, Tao Gui, and Xuanjing Huang.
\newblock Orthogonal subspace learning for language model continual learning.
\newblock \emph{arXiv preprint arXiv:2310.14152}, 2023{\natexlab{a}}.

\bibitem[Wang et~al.(2023{\natexlab{b}})Wang, Li, Wu, Soon, and Zhang]{wang2023finvis}
Ziao Wang, Yuhang Li, Junda Wu, Jaehyeon Soon, and Xiaofeng Zhang.
\newblock Finvis-gpt: A multimodal large language model for financial chart analysis.
\newblock \emph{arXiv preprint arXiv:2308.01430}, 2023{\natexlab{b}}.

\bibitem[Wang et~al.(2022)Wang, Zhang, Lee, Zhang, Sun, Ren, Su, Perot, Dy, and Pfister]{l2p}
Zifeng Wang, Zizhao Zhang, Chen-Yu Lee, Han Zhang, Ruoxi Sun, Xiaoqi Ren, Guolong Su, Vincent Perot, Jennifer Dy, and Tomas Pfister.
\newblock Learning to prompt for continual learning.
\newblock In \emph{Proceedings of the IEEE/CVF conference on computer vision and pattern recognition}, pp.\  139--149, 2022.

\bibitem[Wei et~al.(2025)Wei, Tian, and Li]{wei2025asymlora}
Xuyang Wei, Chunlin Tian, and Li~Li.
\newblock Asymlora: Harmonizing data conflicts and commonalities in mllms.
\newblock \emph{arXiv preprint arXiv:2502.20035}, 2025.

\bibitem[Wu et~al.(2024)Wu, Luo, Li, Pan, Vu, and Haffari]{wu2024continual}
Tongtong Wu, Linhao Luo, Yuan-Fang Li, Shirui Pan, Thuy-Trang Vu, and Gholamreza Haffari.
\newblock Continual learning for large language models: A survey.
\newblock \emph{arXiv preprint arXiv:2402.01364}, 2024.

\bibitem[Yang et~al.(2024)Yang, Chen, Zhang, Liu, Zhang, Ma, Verma, Zhang, Zhou, King, et~al.]{yang2024low}
Menglin Yang, Jialin Chen, Yifei Zhang, Jiahong Liu, Jiasheng Zhang, Qiyao Ma, Harshit Verma, Qianru Zhang, Min Zhou, Irwin King, et~al.
\newblock Low-rank adaptation for foundation models: A comprehensive review.
\newblock \emph{arXiv preprint arXiv:2501.00365}, 2024.

\bibitem[Yin et~al.(2022)Yin, Li, and Xiong]{yin2022contintin}
Wenpeng Yin, Jia Li, and Caiming Xiong.
\newblock Contintin: Continual learning from task instructions.
\newblock \emph{arXiv preprint arXiv:2203.08512}, 2022.

\bibitem[Ying et~al.(2024)Ying, Meng, Wang, Li, Lin, Yang, Zhang, Zhang, Lin, Liu, et~al.]{mmtbench}
Kaining Ying, Fanqing Meng, Jin Wang, Zhiqian Li, Han Lin, Yue Yang, Hao Zhang, Wenbo Zhang, Yuqi Lin, Shuo Liu, et~al.
\newblock Mmt-bench: A comprehensive multimodal benchmark for evaluating large vision-language models towards multitask agi.
\newblock \emph{arXiv preprint arXiv:2404.16006}, 2024.

\bibitem[Zan et~al.(2022)Zan, Chen, Yang, Lin, Kim, Guan, Wang, Chen, and Lou]{zan2022cert}
Daoguang Zan, Bei Chen, Dejian Yang, Zeqi Lin, Minsu Kim, Bei Guan, Yongji Wang, Weizhu Chen, and Jian-Guang Lou.
\newblock Cert: continual pre-training on sketches for library-oriented code generation.
\newblock \emph{arXiv preprint arXiv:2206.06888}, 2022.

\bibitem[Zeng et~al.(2024)Zeng, Zhu, Guo, Zhang, and Liu]{zeng2024modalprompt}
Fanhu Zeng, Fei Zhu, Haiyang Guo, Xu-Yao Zhang, and Cheng-Lin Liu.
\newblock Modalprompt: Dual-modality guided prompt for continual learning of large multimodal models.
\newblock \emph{arXiv preprint arXiv:2410.05849}, 2024.

\bibitem[Zhai et~al.(2023)Zhai, Tong, Li, Cai, Qu, Lee, and Ma]{zhai2023investigating}
Yuexiang Zhai, Shengbang Tong, Xiao Li, Mu~Cai, Qing Qu, Yong~Jae Lee, and Yi~Ma.
\newblock Investigating the catastrophic forgetting in multimodal large language models.
\newblock \emph{arXiv preprint arXiv:2309.10313}, 2023.

\bibitem[Zhang et~al.(2024{\natexlab{a}})Zhang, Lei, Gui, Yang, He, Wang, and Xu]{zhang2024cppo}
Han Zhang, Yu~Lei, Lin Gui, Min Yang, Yulan He, Hui Wang, and Ruifeng Xu.
\newblock Cppo: Continual learning for reinforcement learning with human feedback.
\newblock In \emph{The Twelfth International Conference on Learning Representations}, 2024{\natexlab{a}}.

\bibitem[Zhang et~al.(2024{\natexlab{b}})Zhang, Wei, Jiang, Guo, Li, Zhang, Tong, Liu, Zhou, Wei, et~al.]{zhang2024mavismathematicalvisualinstruction}
Renrui Zhang, Xinyu Wei, Dongzhi Jiang, Ziyu Guo, Shicheng Li, Yichi Zhang, Chengzhuo Tong, Jiaming Liu, Aojun Zhou, Bin Wei, et~al.
\newblock Mavis: Mathematical visual instruction tuning with an automatic data engine.
\newblock \emph{arXiv preprint arXiv:2407.08739}, 2024{\natexlab{b}}.

\end{thebibliography}
\bibliographystyle{iclr2026_conference}

\appendix

\clearpage

\section*{Appendix}
\crefalias{section}{appendix}
\crefalias{subsection}{appendix}

\section{Implementation Details} \label{sec:implement}
In this section, we introduce the implementation details of~\ours~and the evaluation details of each task in domain continual learning and ability continual learning.
\subsection{Training Details}\label{appdx:train}
\textbf{DCL.} \Cref{tab:mrlora:dcl:hyperparams} shows the hyperparameters for training the router and expert in domain continual learning.
For most configurations, we follow the default setting of LLaVA 1.5~\citep{liu2023llava}.
To ensure comparable training exposure across datasets of varying sizes, each task is trained for approximately 60,000 instances in DCL. For efficient fine-tuning, a rank of 32 is employed.
For all the experiments, we use 8 A100 GPUs, and the training time for each task is around 1 hour.

\textbf{ACL.} \Cref{tab:mrlora:acl:hyperparams} shows the hyperparameters for ability continual learning.
For ability continual learning, training time is around 20 hours to train all the tasks sequentially.

\textbf{Router Training.} For the router training, we train 30 epochs in domain continual learning and ability continual learning; we keep other configurations identical to the experts' except for the learning rate.
We use the codebase from MCITlib~\citep{guo2025mcitlib} and LLaVA~\citep{liu2023llava}.

\begin{table}[h]
\caption{Hyperparameters of MR-LoRA in domain continual learning}
\label{tab:mrlora:dcl:hyperparams}

\centering
\begin{tabular}{l|c|c|c|c}
\toprule
\multicolumn{1}{l}{}    & \multicolumn{2}{c}{Expert Config} & \multicolumn{2}{c}{Router Config} \\ \cmidrule(lr){2-3} \cmidrule(lr){4-5}
                        & \multicolumn{1}{c|}{LLaVA} & InternVL & LLaVA          & InternVL     \\ 
\midrule \midrule
optimizer               & \multicolumn{2}{c|}{AdamW}        & \multicolumn{2}{c}{AdamW}         \\
batch size              & \multicolumn{2}{c|}{$64$}         & \multicolumn{2}{c}{$64$}          \\
lr schedule             & \multicolumn{2}{c|}{cosine decay} & \multicolumn{2}{c}{cosine decay}  \\
lr warmup ratio         & \multicolumn{2}{c|}{$0.03$}       & \multicolumn{2}{c}{$0.03$}        \\
LoRA rank               & \multicolumn{2}{c|}{$32$}         & \multicolumn{2}{c}{$32$}          \\
DeepSpeed stage         & \multicolumn{2}{c|}{$2$}          & \multicolumn{2}{c}{$2$}           \\
\midrule
base lr                 & \multicolumn{2}{c|}{\num{1e-4}}   & \num{2e-5}    & \num{1e-4}        \\
\midrule
epoch for RS            & \multicolumn{2}{c|}{$1$}          & \multicolumn{2}{c}{-}             \\
epoch for Med           & \multicolumn{2}{c|}{$3$}          & \multicolumn{2}{c}{$30$}          \\
epoch for AD            & \multicolumn{2}{c|}{$1$}          & \multicolumn{2}{c}{$30$}          \\
epoch for Sci           & \multicolumn{2}{c|}{$2$}          & \multicolumn{2}{c}{$30$}          \\
epoch for Fin           & \multicolumn{2}{c|}{$1$}          & \multicolumn{2}{c}{$30$}          \\
\bottomrule
\end{tabular}

\end{table}

\subsection{Evaluation Details}
In domain continual learning, for the financial task, all the questions are MCQ or Y/N questions; we require the prediction to exactly match the ground truth.
For autonomous driving, medical, and remote sensing tasks, we consider the prediction to include the ground truth as the correct answer. This serves as the default evaluation method.
For science tasks, some test samples are multiple-choice questions (MCQs), and predictions are required to exactly match the ground truth. Certain questions in MapQA~\citep{chang2022mapqa} require the model to list places; in these cases, we compute the percentage of correct responses. Other science questions are evaluated according to the default method.
In ability continual learning, we follow the default setting of the corresponding benchmarks.

\subsection{Detailed Evalution Metrics} \label{appdx:eval:metrics}
We used the integrated metrics in SEFE and MCITlib~\citep{chen2025sefe,guo2025mcitlib} to evaluate the performance of each method.
\begin{itemize}
\item \textbf{Last} accuracy is the accuracy of all seen tasks after learning the last task.
    \item \textbf{Mean Finetune Accuracy (MFT)} measures the average accuracy achieved on each task immediately after it is learned, serving as an upper bound that reflects the model’s performance in the absence of forgetting.
    \item \textbf{Mean Final Accuracy (MFN)} computes the average accuracy over all tasks after completing the full incremental training process, representing the model’s overall retained performance.
    \item \textbf{Mean Average Accuracy (MAA)} calculates the mean of average accuracies on all learned tasks after each training step, offering a holistic view of performance throughout the continual learning process.
    \item \textbf{Backward Transfer (BWT)} captures the change in accuracy for each task by comparing its final accuracy with that immediately after it was learned, quantifying the extent of forgetting.
\end{itemize}

\begin{figure}[ht]
    \centering
    \includegraphics[width=0.6\linewidth]{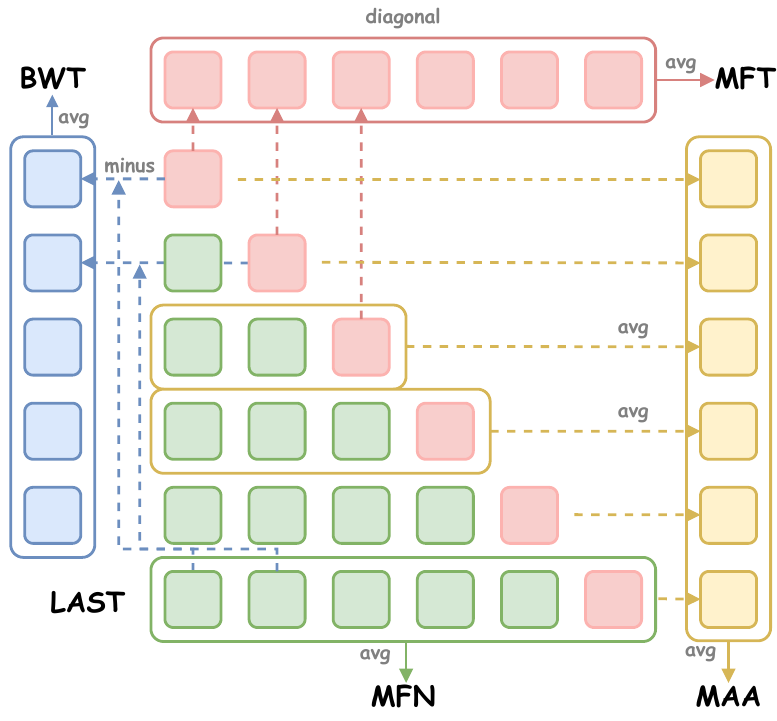}
    \caption{Illustration of the evaluation metric calculations~\citep{guo2025mcitlib}.}
    \label{fig:evaluation}
\end{figure}

For clarity, a conceptual illustration of the evaluation metrics is provided in ~\Cref{fig:evaluation}.

\begin{table}[t]
\caption{Hyperparameters of MR-LoRA in ability continual learning}
\label{tab:mrlora:acl:hyperparams}

\centering
\begin{tabular}{l|c|c|c|c}
\toprule
\multicolumn{1}{l}{}    & \multicolumn{2}{c}{Expert Config} & \multicolumn{2}{c}{Router Config} \\ \cmidrule(lr){2-3} \cmidrule(lr){4-5}
                        & \multicolumn{1}{c|}{LLaVA} & InternVL & LLaVA          & InternVL     \\ 
\midrule \midrule
optimizer               & \multicolumn{2}{c|}{AdamW}        & \multicolumn{2}{c}{AdamW}         \\
batch size              & \multicolumn{2}{c|}{$128$}        & \multicolumn{2}{c}{$128$}         \\
lr schedule             & \multicolumn{2}{c|}{cosine decay} & \multicolumn{2}{c}{cosine decay}  \\
lr warmup ratio         & \multicolumn{2}{c|}{$0.03$}       & \multicolumn{2}{c}{$0.03$}        \\
LoRA rank               & \multicolumn{2}{c|}{$32$}         & \multicolumn{2}{c}{$32$}          \\
DeepSpeed stage         & \multicolumn{2}{c|}{$2$}          & \multicolumn{2}{c}{$2$}           \\
\midrule
base lr (OCR)           & \num{5e-5}  & \num{2e-4}          & -           & -                   \\
base lr (M\&L,VP,GUI)   & \num{2e-4}  & \num{2e-4}          & \num{2e-4}  & \num{1e-4}          \\
\midrule
epoch for OCR           & \multicolumn{2}{c|}{$3$}          & \multicolumn{2}{c}{-}             \\
epoch for Math \& Logic & \multicolumn{2}{c|}{$1$}          & \multicolumn{2}{c}{$30$}          \\
epoch for VP            & \multicolumn{2}{c|}{$1$}          & \multicolumn{2}{c}{$30$}          \\
epoch for GUI Agent     & \multicolumn{2}{c|}{$3$}          & \multicolumn{2}{c}{$30$}          \\
\bottomrule
\end{tabular}

\end{table}

\subsection{Router prompt for MR-LoRA} \label{appdx:prompt:acl:router}
\begin{figure}[H]
    \centering
    \includegraphics[width=\linewidth]{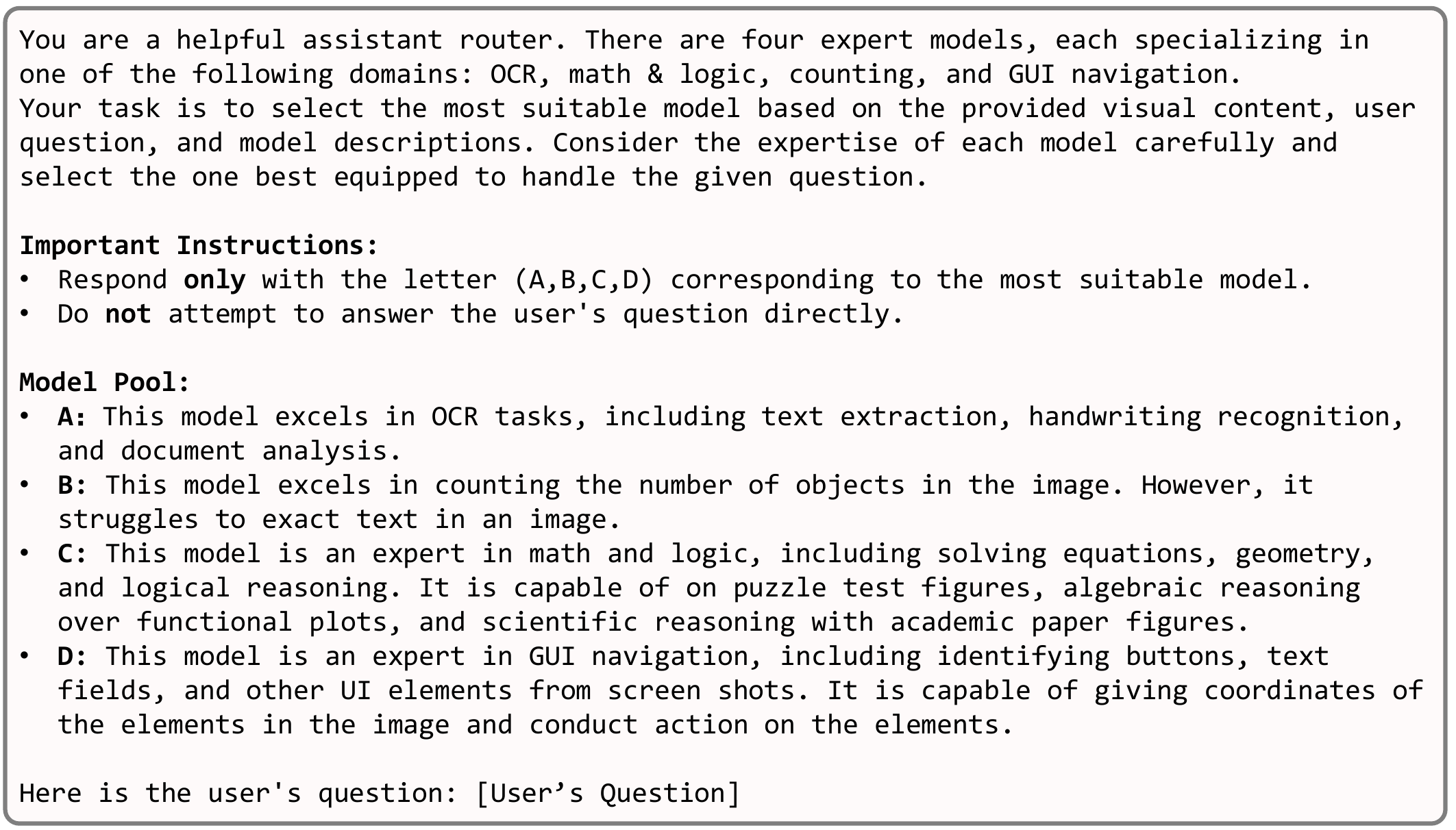}
    \caption{Prompt for the router in ability continual learning.}
    \label{fig:prompt:acl:router}
\end{figure}

We previously provided our router prompt for DCL in \Cref{fig:router_instruct}. The prompt for ACL appears in \Cref{fig:prompt:acl:router}.

\begin{figure}[ht]
    \centering
    \includegraphics[width=\linewidth]{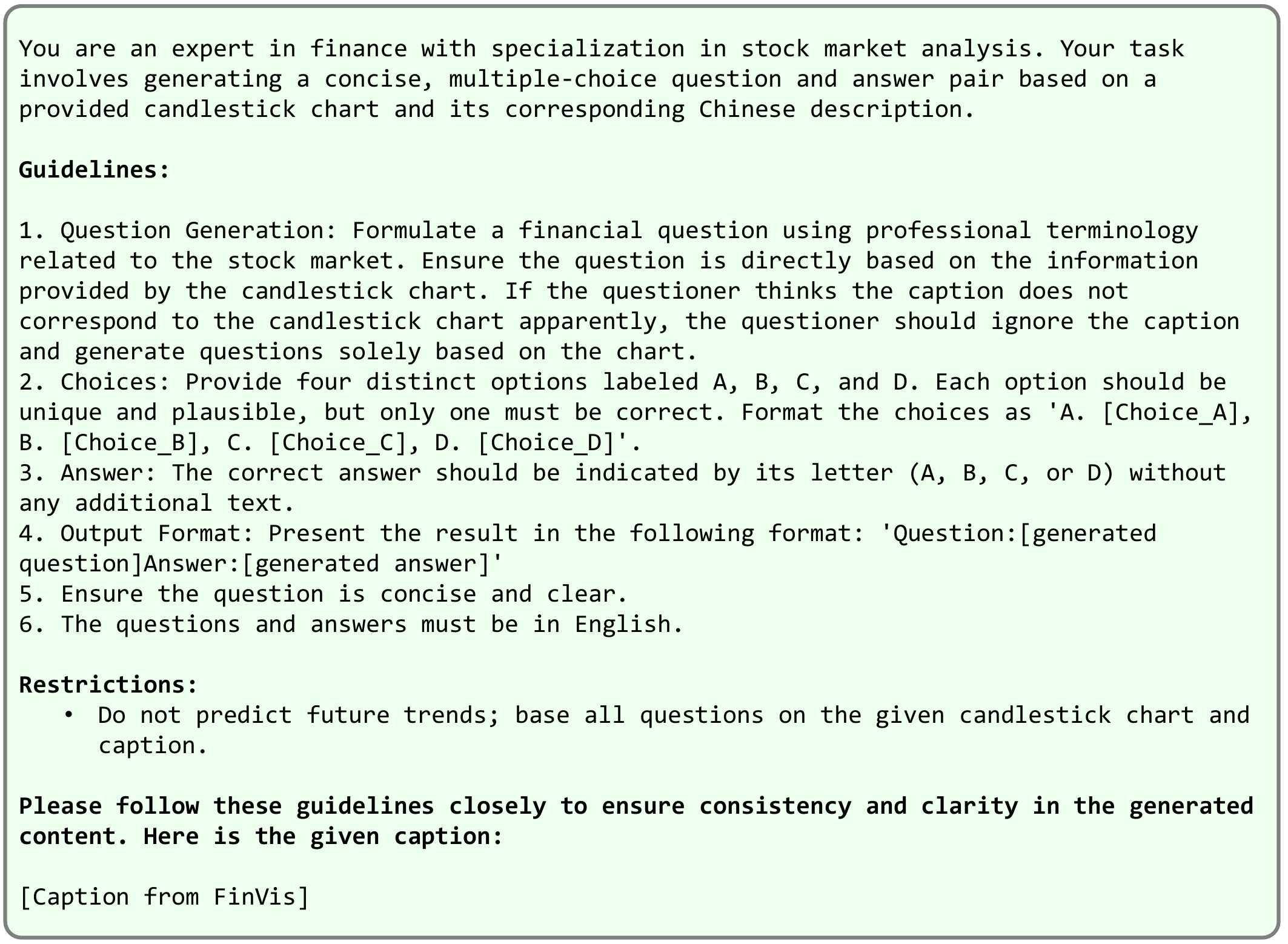}
    \caption{Prompt for the Questioner to generate MCQ question answer pairs.}
    \label{fig:MCQ}
\end{figure}
\begin{figure}[ht]
    \centering
    \includegraphics[width=\linewidth]{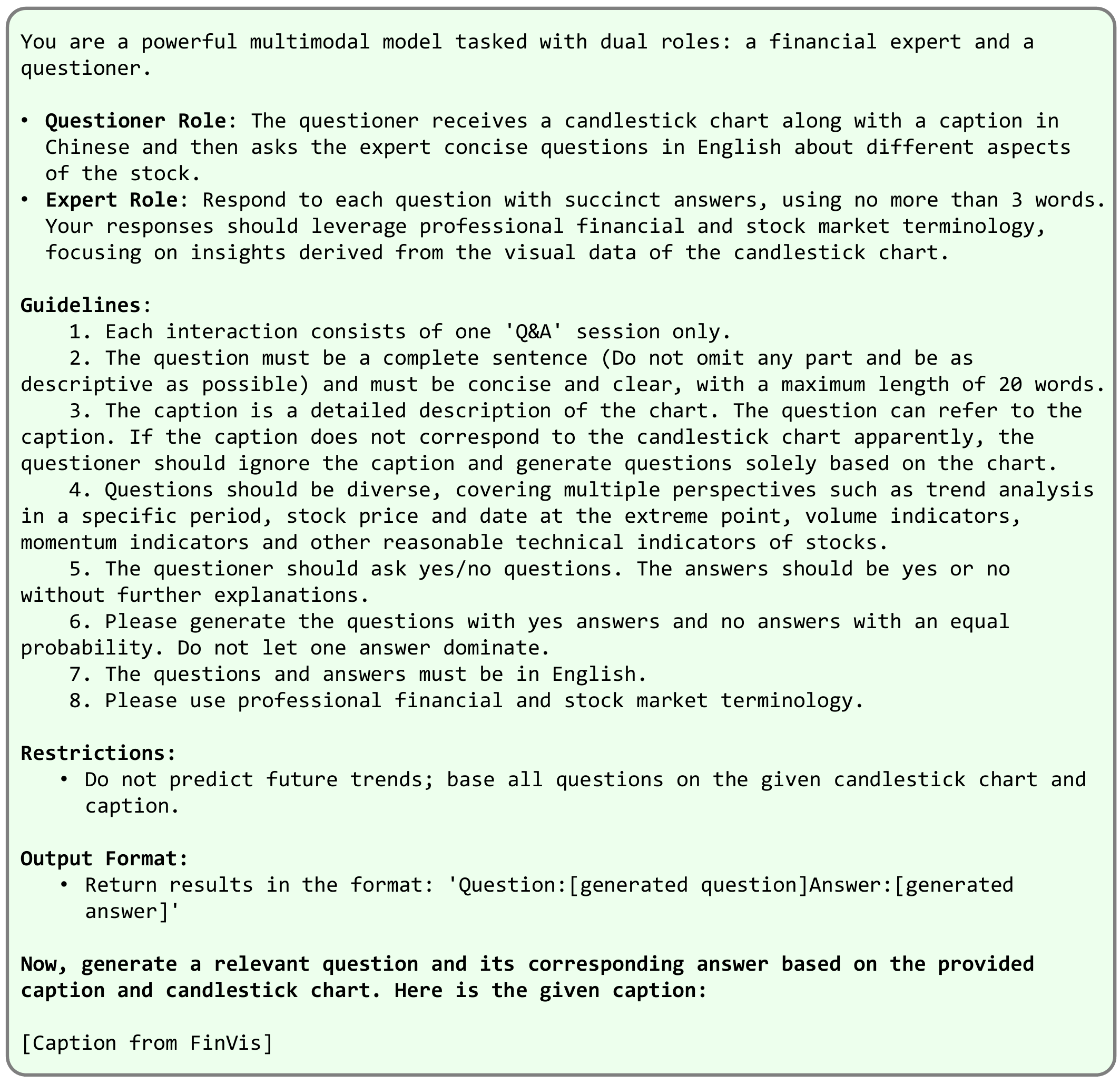}
    \caption{Prompt for the Questioner to generate Y/N question answer pairs.}
    \label{fig:PromptTF}
\end{figure}
\begin{figure}[ht]
    \centering
    \includegraphics[width=\linewidth]{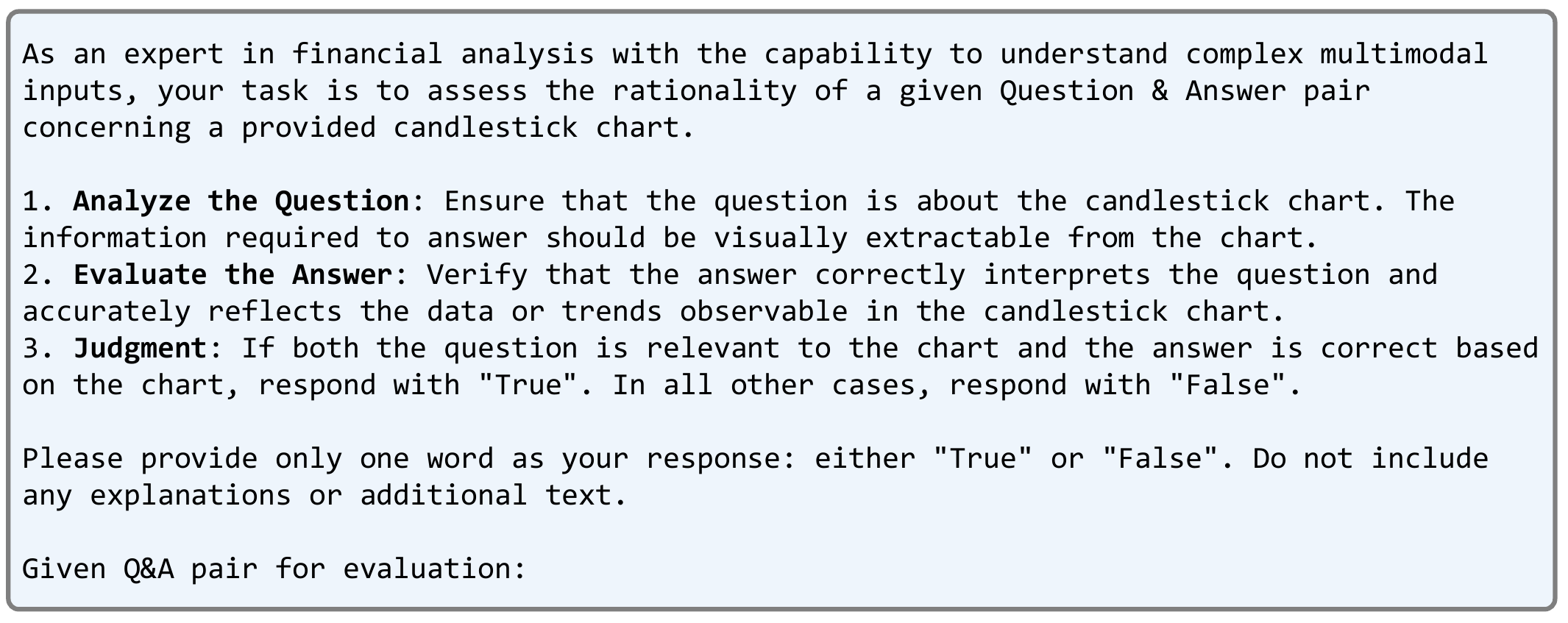}
    \caption{Prompt for the Inspector to check the question answer pairs.}
    \label{fig:check}
\end{figure}

\section{Details of StockQA Dataset} \label{appdx:stockqa}
\textbf{Overview.}
The StockQA dataset is a multimodal financial dataset concentrated on stock analysis.
It is rewritten from the FinVis~\citep{wang2023finvis} dataset.

Finvis dataset is a Chinese caption dataset generated by GPT4V~\citep{achiam2023gpt}.
All the captions are related to the stock technical indicator analysis.
However, the caption form is not convenient for evaluation, and there may be a language gap between this task and other tasks.
Therefore, we use a \textit{questioner-inspector} data pipeline with a powerful MLLM Qwen2.5-VL~\citep{Qwen2.5-VL} to rewrite the caption into MCQ and Y/N question-answer pairs and name it StockQA.
When manually checking the inspector process, we find that the inspector \textit{misclassified} some correct question-answer pairs.
Nevertheless, it successfully identified erroneous instances, thereby contributing to the overall correctness of the final dataset.

\textbf{Prompts for agents.}
\Cref{fig:PromptTF,fig:MCQ} shows the prompt we use for the Questioner to generate Y/N and MCQ question-answer pairs, respectively.
\Cref{fig:check} is the prompt we use for the inspector.

\textbf{Rules for filtering.}
After using an inspector agent to check the correctness and rationality, we employ the following rules to balance the choices of multiple choice questions to mitigate the position bias~\citep{liu2024lost} and format the output.
\begin{itemize}
    \item \textbf{Format:} Remove the unnecessary spaces, line breaks, and punctuation to make each question in the same format.
    \item \textbf{Position:} Exchange the choices of multiple choice questions to ensure the right answers of the total datasets are distributed with the same probability.
\end{itemize}

\begin{table}[H]
    \centering\small
    \caption{Statistics of the StockQA dataset.}
    \vspace{3pt}
    \begin{tabular}{c|cccc}
    \toprule
        Data & Max Length & Min Length & Average Length & Amount\\
        \midrule 
         MCQ& 683& 115&392.74&48k\\
         TF&99&21&42.29&22k\\
         Total&683&21&282.60&70k \\
         \bottomrule
    \end{tabular}
    \label{tab:stock}
\end{table}
\begin{table}[H]
    \centering\small
    \setlength{\tabcolsep}{7pt}\caption{Existing assets grouped by license.}
    \begin{tabular}{ll}
    \toprule
    License & Assets \\ 
    \midrule
    CC-BY-SA-4.0 & TQA, MapQA, MathVista, AI2D \\ 
    Apache-2.0 & DriveLM, MathV360k, CV-Bench, CoIN \\ 
    MIT & Monkey, OCRbench, MAVIS \\ 
    CC-BY-4.0 & CLEVR, ScreenQA, Screen2Words, MMTBench \\ 
    \bottomrule
    \end{tabular}
    \label{tab:license_reorganized}
\end{table}
\begin{figure}[!t]
    \centering
    \includegraphics[width=0.8\linewidth]{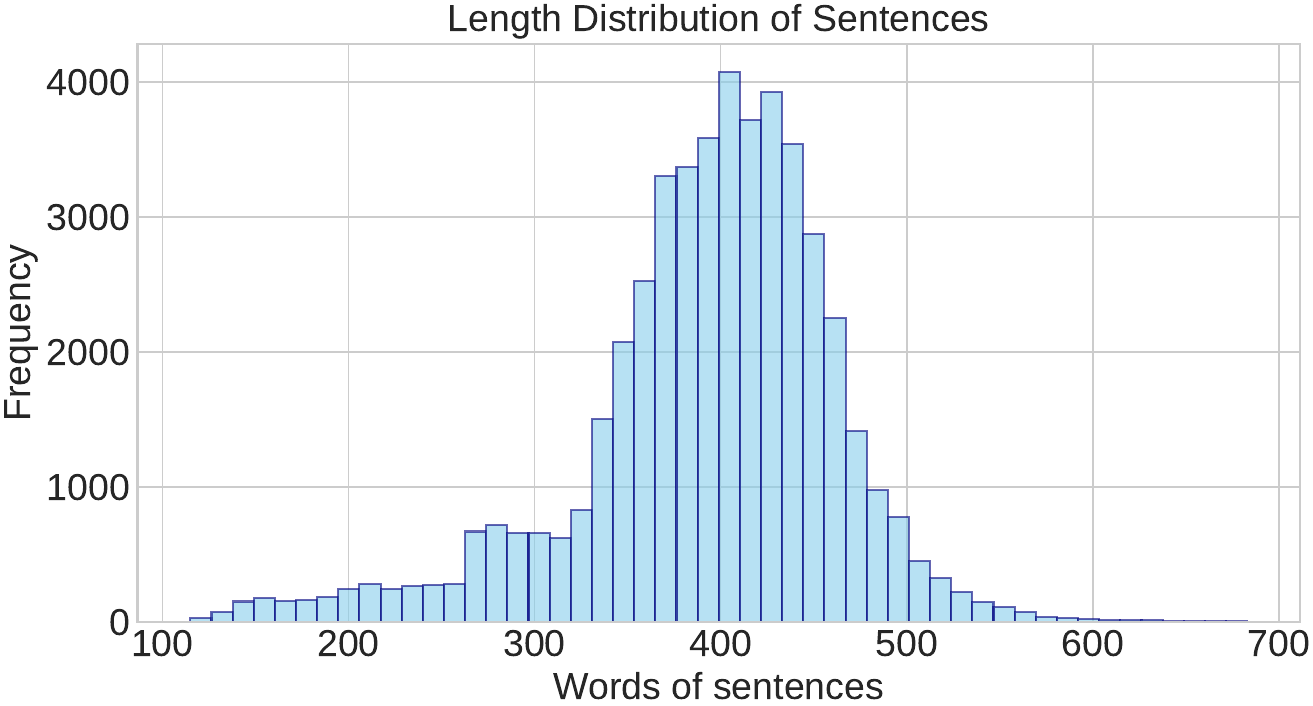}
    \caption{Word length distribution of the StockQA dataset.}
    \label{fig:length}
\end{figure}

\begin{figure}[!t]
    \centering
    \includegraphics[width=\linewidth]{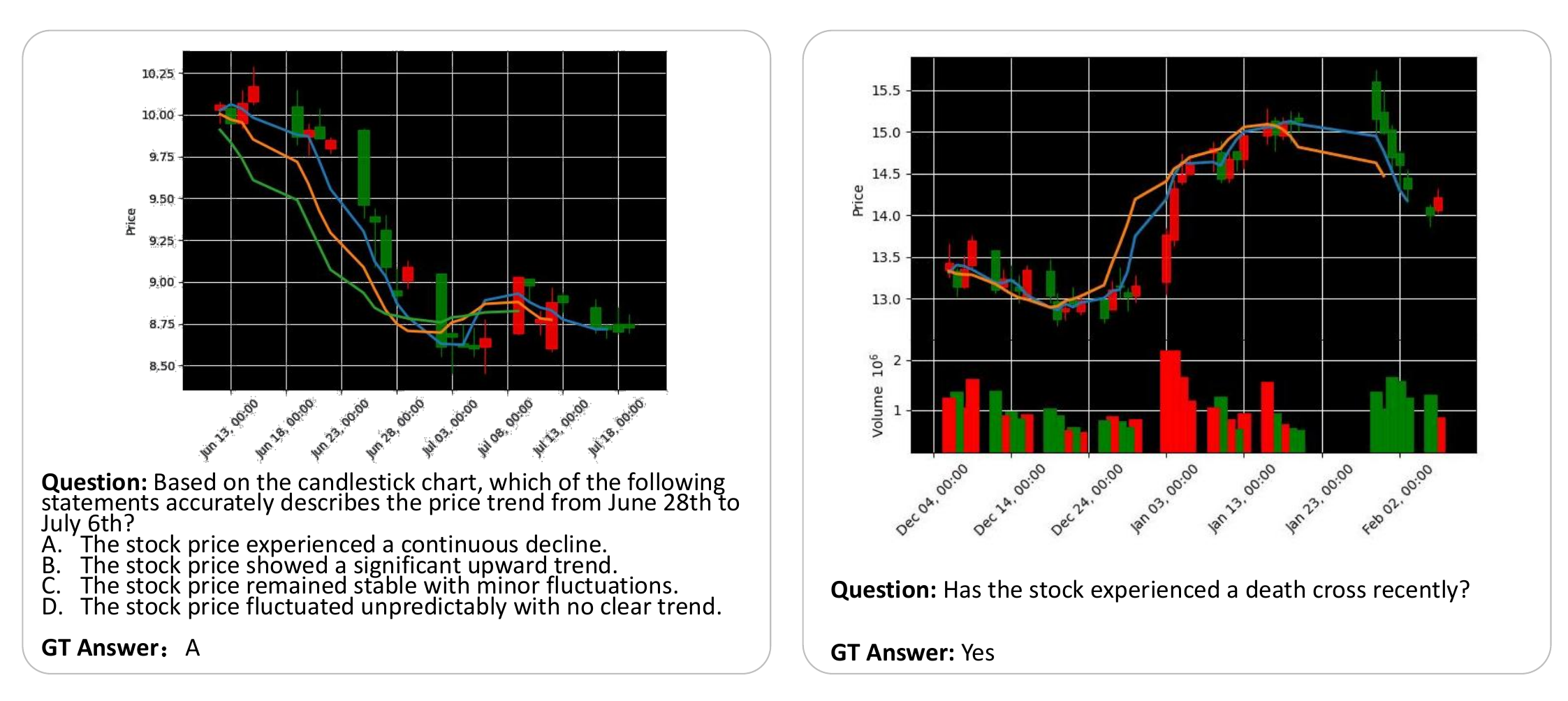}
    \caption{MCQ and Y/N examples in StockQA dataset.}
    \label{fig:stock_example}
\end{figure}

\begin{figure}[ht]
    \centering
    \includegraphics[width=0.9\linewidth]{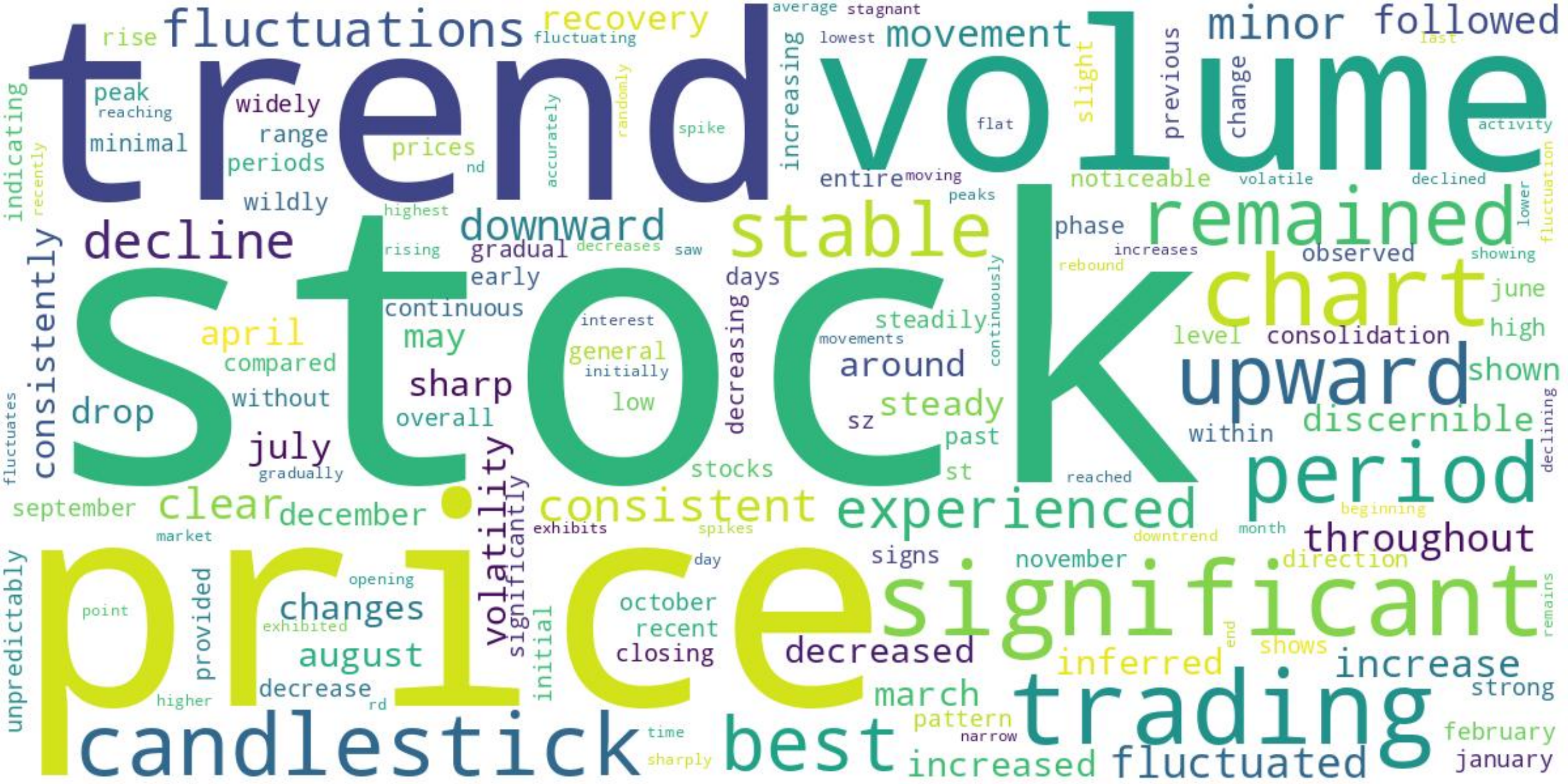}
    \caption{Word cloud of StockQA dataset.}
    \label{fig:cloud}
\end{figure}

\textbf{Statistics of StockQA dataset.}
StockQA is a new VQA dataset related to multimodal stock analysis.
It includes 70k question-answer pairs. of which 60k is the training set and 10k is the test set.
For the training data, there are 40k MCQ and 20k Y/N QA pairs. For the test data, there are 8k MCQ and 2k QA pairs.
Each choice is equally distributed after our cleaning process.
\Cref{fig:cloud,fig:stock_example} shows the word cloud and examples of StockQA dataset.
\Cref{tab:stock,fig:length} shows the detailed statistics of StockQA dataset.

\textbf{Dataset License.}
Our dataset follows the CC-BY license. This license allows reusers to distribute, remix, adapt, and build upon the material in any medium or format, so long as attribution is given to the creator. The license allows for commercial use.
For other assets we used, we list the licenses below in  \Cref{tab:license_reorganized}.

\clearpage

\section{Detailed Continual Learning Results} \label{sec:bigmatrix}
In this section, we show the detailed inference results of all the methods (LoRA~\citep{hu2021lora}, LoRA$^*$~\citep{hu2021lora}, 
O-LoRA~\citep{wang2023orthogonal}, O-LoRA$^*$~\citep{wang2023orthogonal}, 
MoELoRA~\citep{chen2024coin}, MoELoRA$^*$~\citep{chen2024coin}, 
CL-MoE~\citep{huai2025cl}, CL-MoE$^*$~\citep{huai2025cl}, 
HiDe~\citep{guo2025hide}, HiDe$^*$~\citep{guo2025hide},
SEFE~\citep{chen2025sefe}, SEFE$^*$~\citep{chen2025sefe}, 
DISCO~\citep{guo2025federated}, DISCO$^*$~\citep{guo2025federated}
and~\ours) during each continual learning stage, where~$^*$ denotes the original method with replay data.

\subsection{Baseline Results in Domain Continual Learning}

\begin{table}[H]
\centering \setlength{\tabcolsep}{4pt} \small
\caption{Result matrices of InternVL-based baselines in domain continual learning. $^{*}$ denotes
the original method with replay data.}
\label{tab:dclresults:internvl}

\begin{subtable}[t]{0.49\linewidth}\centering
\begin{tabular}{Lccccc}
\toprule
LoRA-FT & RS & Med & AD & Sci & Fin \\
\midrule
RS & 81.29 &  &  &  &  \\
Med & 75.71 & 65.92 &  &  &  \\
AD & 69.38 & 56.87 & 53.56 &  &  \\
Sci & 71.12 & 53.75 & 46.83 & 53.48 &  \\
Fin & 69.93 & 52.17 & 33.04 & 42.67 & 91.07 \\
\bottomrule
\end{tabular}

\end{subtable}\hfill
\begin{subtable}[t]{0.49\linewidth}\centering
\begin{tabular}{Lccccc}
\toprule
LoRA-FT$^*$ & RS & Med & AD & Sci & Fin \\
\midrule
RS & 81.68 &  &  &  &  \\
Med & 77.45 & 66.69 &  &  &  \\
AD & 77.24 & 61.32 & 53.81 &  &  \\
Sci & 77.89 & 55.43 & 49.13 & 53.53 &  \\
Fin & 77.06 & 47.55 & 42.67 & 43.31 & 91.44 \\
\bottomrule
\end{tabular}

\end{subtable}\\[8pt]
\begin{subtable}[t]{0.49\linewidth}\centering
\begin{tabular}{Lccccc}
\toprule
MoELoRA & RS & Med & AD & Sci & Fin \\
\midrule
RS & 81.22 &  &  &  &  \\
Med & 77.56 & 66.00 &  &  &  \\
AD & 74.56 & 58.74 & 53.62 &  &  \\
Sci & 72.62 & 54.77 & 47.65 & 52.75 &  \\
Fin & 69.90 & 52.08 & 33.17 & 42.19 & 90.58 \\
\bottomrule
\end{tabular}

\end{subtable}\hfill
\begin{subtable}[t]{0.49\linewidth}\centering
\begin{tabular}{Lccccc}
\toprule
MoELoRA$^*$ & RS & Med & AD & Sci & Fin \\
\midrule
RS & 80.75 &  &  &  &  \\
Med & 78.10 & 64.77 &  &  &  \\
AD & 73.24 & 59.54 & 52.90 &  &  \\
Sci & 76.82 & 53.64 & 42.11 & 51.24 &  \\
Fin & 76.74 & 52.65 & 38.81 & 42.15 & 89.84 \\
\bottomrule
\end{tabular}

\end{subtable}\\[8pt]
\begin{subtable}[t]{0.49\linewidth}\centering
\begin{tabular}{Lccccc}
\toprule
HiDe & RS & Med & AD & Sci & Fin \\
\midrule
RS & 81.24 &  &  &  &  \\
Med & 79.59 & 64.71 &  &  &  \\
AD & 78.85 & 58.37 & 41.75 &  &  \\
Sci & 78.33 & 58.51 & 39.94 & 49.99 &  \\
Fin & 75.40 & 57.66 & 36.73 & 41.48 & 88.59 \\
\bottomrule
\end{tabular}

\end{subtable}\hfill
\begin{subtable}[t]{0.49\linewidth}\centering
\begin{tabular}{Lccccc}
\toprule
HiDe$^*$ & RS & Med & AD & Sci & Fin \\
\midrule
RS & 73.92 &  &  &  &  \\
Med & 71.44 & 64.22 &  &  &  \\
AD & 65.74 & 49.83 & 42.92 &  &  \\
Sci & 70.64 & 54.79 & 40.47 & 50.75 &  \\
Fin & 53.17 & 52.61 & 40.85 & 47.04 & 89.17 \\
\bottomrule
\end{tabular}

\end{subtable}\\[8pt]
\begin{subtable}[t]{0.49\linewidth}\centering
\begin{tabular}{Lccccc}
\toprule
DISCO & RS & Med & AD & Sci & Fin \\
\midrule
RS & 81.42 &  &  &  &  \\
Med & 79.13 & 63.80 &  &  &  \\
AD & 78.62 & 60.79 & 53.98 &  &  \\
Sci & 77.40 & 52.21 & 53.74 & 54.18 &  \\
Fin & 75.12 & 50.69 & 52.41 & 50.67 & 90.86 \\
\bottomrule
\end{tabular}

\end{subtable}\hfill
\begin{subtable}[t]{0.49\linewidth}\centering
\begin{tabular}{Lccccc}
\toprule
DISCO$^*$ & RS & Med & AD & Sci & Fin \\
\midrule
RS & 81.49 &  &  &  &  \\
Med & 80.14 & 63.05 &  &  &  \\
AD & 78.87 & 57.42 & 53.77 &  &  \\
Sci & 78.67 & 52.80 & 53.56 & 53.52 &  \\
Fin & 77.90 & 47.50 & 49.13 & 49.37 & 90.92 \\
\bottomrule
\end{tabular}

\end{subtable}\\[8pt]
\end{table}

\newpage
\begin{table}[H]
\centering \setlength{\tabcolsep}{4pt} \small
\caption{Result matrices of LLaVA-based baselines in domain continual learning. $^{*}$ denotes
the original method with replay data.}
\label{tab:dclresults:llava}

\begin{subtable}[t]{0.49\linewidth}\centering
\begin{tabular}{Lccccc}
\toprule
LoRA-FT & RS & Med & AD & Sci & Fin \\
\midrule
RS & 78.32 &  &  &  &  \\
Med & 74.68 & 57.53 &  &  &  \\
AD & 68.93 & 47.19 & 52.15 &  &  \\
Sci & 75.12 & 45.56 & 38.46 & 49.44 &  \\
Fin & 69.65 & 41.59 & 25.43 & 40.88 & 87.45 \\
\bottomrule
\end{tabular}

\end{subtable}\hfill
\begin{subtable}[t]{0.49\linewidth}\centering
\begin{tabular}{Lccccc}
\toprule
LoRA-FT$^*$ & RS & Med & AD & Sci & Fin \\
\midrule
RS & 79.33 &  &  &  &  \\
Med & 76.45 & 57.58 &  &  &  \\
AD & 74.54 & 54.26 & 52.96 &  &  \\
Sci & 77.00 & 50.31 & 45.13 & 51.88 &  \\
Fin & 76.54 & 50.27 & 43.01 & 43.32 & 89.85 \\
\bottomrule
\end{tabular}

\end{subtable}\\[8pt]
\begin{subtable}[t]{0.49\linewidth}\centering
\begin{tabular}{Lccccc}
\toprule
O-LoRA & RS & Med & AD & Sci & Fin \\
\midrule
RS & 79.25 &  &  &  &  \\
Med & 74.05 & 56.52 &  &  &  \\
AD & 76.06 & 43.71 & 52.32 &  &  \\
Sci & 76.60 & 44.87 & 40.57 & 50.58 &  \\
Fin & 74.64 & 44.42 & 30.02 & 41.47 & 87.15 \\
\bottomrule
\end{tabular}

\end{subtable}\hfill
\begin{subtable}[t]{0.49\linewidth}\centering
\begin{tabular}{Lccccc}
\toprule
O-LoRA$^*$ & RS & Med & AD & Sci & Fin \\
\midrule
RS & 79.17 &  &  &  &  \\
Med & 78.21 & 56.65 &  &  &  \\
AD & 77.52 & 38.60 & 37.81 &  &  \\
Sci & 77.61 & 44.22 & 35.40 & 45.59 &  \\
Fin & 76.94 & 41.17 & 34.18 & 39.61 & 83.22 \\
\bottomrule
\end{tabular}

\end{subtable}\\[8pt]
\begin{subtable}[t]{0.49\linewidth}\centering
\begin{tabular}{Lccccc}
\toprule
MoELoRA & RS & Med & AD & Sci & Fin \\
\midrule
RS & 79.09 &  &  &  &  \\
Med & 74.78 & 58.73 &  &  &  \\
AD & 77.69 & 43.72 & 51.47 &  &  \\
Sci & 76.87 & 43.79 & 32.81 & 48.67 &  \\
Fin & 77.54 & 41.85 & 27.62 & 40.13 & 86.75 \\
\bottomrule
\end{tabular}

\end{subtable}\hfill
\begin{subtable}[t]{0.49\linewidth}\centering
\begin{tabular}{Lccccc}
\toprule
MoELoRA$^*$ & RS & Med & AD & Sci & Fin \\
\midrule
RS & 79.66 &  &  &  &  \\
Med & 78.44 & 60.50 &  &  &  \\
AD & 78.54 & 49.86 & 52.54 &  &  \\
Sci & 78.00 & 50.53 & 43.32 & 49.30 &  \\
Fin & 77.63 & 49.54 & 39.08 & 41.04 & 89.21 \\
\bottomrule
\end{tabular}

\end{subtable}\\[8pt]
\begin{subtable}[t]{0.49\linewidth}\centering
\begin{tabular}{Lccccc}
\toprule
CL-MoE & RS & Med & AD & Sci & Fin \\
\midrule
RS & 79.08 &  &  &  &  \\
Med & 73.48 & 60.56 &  &  &  \\
AD & 72.61 & 44.42 & 51.62 &  &  \\
Sci & 71.02 & 48.04 & 37.70 & 50.28 &  \\
Fin & 71.34 & 46.84 & 26.33 & 41.17 & 88.74 \\
\bottomrule
\end{tabular}

\end{subtable}\hfill
\begin{subtable}[t]{0.49\linewidth}\centering
\begin{tabular}{Lccccc}
\toprule
CL-MoE$^*$ & RS & Med & AD & Sci & Fin \\
\midrule
RS & 79.40 &  &  &  &  \\
Med & 76.32 & 61.10 &  &  &  \\
AD & 72.01 & 54.49 & 52.56 &  &  \\
Sci & 76.64 & 53.89 & 43.83 & 49.98 &  \\
Fin & 76.58 & 52.31 & 39.65 & 45.64 & 90.21 \\
\bottomrule
\end{tabular}

\end{subtable}\\[8pt]
\begin{subtable}[t]{0.49\linewidth}\centering
\begin{tabular}{Lccccc}
\toprule
HiDe & RS & Med & AD & Sci & Fin \\
\midrule
RS & 78.14 &  &  &  &  \\
Med & 74.26 & 58.05 &  &  &  \\
AD & 74.90 & 42.94 & 39.65 &  &  \\
Sci & 75.43 & 44.91 & 38.33 & 46.44 &  \\
Fin & 74.31 & 48.95 & 33.21 & 38.54 & 81.55 \\
\bottomrule
\end{tabular}

\end{subtable}\hfill
\begin{subtable}[t]{0.49\linewidth}\centering
\begin{tabular}{Lccccc}
\toprule
HiDe$^*$ & RS & Med & AD & Sci & Fin \\
\midrule
RS & 79.21 &  &  &  &  \\
Med & 77.79 & 60.88 &  &  &  \\
AD & 77.64 & 48.19 & 38.12 &  &  \\
Sci & 77.51 & 48.84 & 35.76 & 46.71 &  \\
Fin & 74.80 & 42.29 & 34.03 & 38.01 & 79.22 \\
\bottomrule
\end{tabular}

\end{subtable}\\[8pt]
\begin{subtable}[t]{0.49\linewidth}\centering
\begin{tabular}{Lccccc}
\toprule
SEFE & RS & Med & AD & Sci & Fin \\
\midrule
RS & 78.27 &  &  &  &  \\
Med & 76.32 & 58.42 &  &  &  \\
AD & 77.22 & 49.13 & 52.49 &  &  \\
Sci & 77.83 & 47.70 & 43.01 & 49.04 &  \\
Fin & 77.26 & 50.37 & 37.21 & 40.87 & 86.82 \\
\bottomrule
\end{tabular}

\end{subtable}\hfill
\begin{subtable}[t]{0.49\linewidth}\centering
\begin{tabular}{Lccccc}
\toprule
SEFE$^*$ & RS & Med & AD & Sci & Fin \\
\midrule
RS & 79.21 &  &  &  &  \\
Med & 78.39 & 60.93 &  &  &  \\
AD & 79.00 & 57.68 & 53.11 &  &  \\
Sci & 78.76 & 51.39 & 47.99 & 51.87 &  \\
Fin & 78.43 & 52.85 & 46.21 & 47.76 & 89.33 \\
\bottomrule
\end{tabular}

\end{subtable}\\[8pt]
\begin{subtable}[t]{0.49\linewidth}\centering
\begin{tabular}{Lccccc}
\toprule
DISCO & RS & Med & AD & Sci & Fin \\
\midrule
RS & 78.57 &  &  &  &  \\
Med & 75.80 & 52.36 &  &  &  \\
AD & 76.37 & 49.78 & 53.04 &  &  \\
Sci & 76.11 & 45.76 & 49.26 & 49.23 &  \\
Fin & 76.03 & 45.20 & 43.79 & 42.33 & 88.95 \\
\bottomrule
\end{tabular}

\end{subtable}\hfill
\begin{subtable}[t]{0.49\linewidth}\centering
\begin{tabular}{Lccccc}
\toprule
DISCO$^*$ & RS & Med & AD & Sci & Fin \\
\midrule
RS & 79.20 &  &  &  &  \\
Med & 77.96 & 52.44 &  &  &  \\
AD & 78.05 & 49.85 & 53.03 &  &  \\
Sci & 77.26 & 46.32 & 53.08 & 51.99 &  \\
Fin & 77.78 & 46.25 & 50.45 & 49.51 & 89.71 \\
\bottomrule
\end{tabular}

\end{subtable}\\[8pt]
\end{table}

\subsection{Baseline Results in Ability Continual Learning}

\begin{table}[H]
\centering \setlength{\tabcolsep}{4pt} \small
\caption{Result matrices of LLaVA-based baselines in ability continual learning. $^{*}$ denotes
the original method with replay data.}
\label{tab:aclresults:llava}

\begin{subtable}[t]{0.49\linewidth}\centering
\begin{tabular}{Lcccc}
\toprule
LoRA-FT & OCR & Math & VP & APP \\
\midrule
OCR & 33.30 &  &  &  \\
Math & 32.60 & 34.20 &  &  \\
VP & 31.70 & 32.80 & 65.10 &  \\
APP & 23.60 & 33.70 & 55.84 & 32.50 \\
\bottomrule
\end{tabular}

\end{subtable}\hfill
\begin{subtable}[t]{0.49\linewidth}\centering
\begin{tabular}{Lcccc}
\toprule
LoRA-FT$^*$ & OCR & Math & VP & APP \\
\midrule
OCR & 32.60 &  &  &  \\
Math & 33.60 & 33.80 &  &  \\
VP & 31.10 & 33.50 & 66.12 &  \\
APP & 21.80 & 32.70 & 58.38 & 28.75 \\
\bottomrule
\end{tabular}

\end{subtable}\\[8pt]
\begin{subtable}[t]{0.49\linewidth}\centering
\begin{tabular}{Lcccc}
\toprule
O-LoRA & OCR & Math & VP & APP \\
\midrule
OCR & 32.90 &  &  &  \\
Math & 29.80 & 33.60 &  &  \\
VP & 27.40 & 33.70 & 58.63 &  \\
APP & 29.60 & 32.90 & 52.41 & 33.75 \\
\bottomrule
\end{tabular}

\end{subtable}\hfill
\begin{subtable}[t]{0.49\linewidth}\centering
\begin{tabular}{Lcccc}
\toprule
O-LoRA$^*$ & OCR & Math & VP & APP \\
\midrule
OCR & 34.00 &  &  &  \\
Math & 28.40 & 36.80 &  &  \\
VP & 28.90 & 33.90 & 61.55 &  \\
APP & 29.60 & 31.30 & 60.79 & 27.50 \\
\bottomrule
\end{tabular}

\end{subtable}\\[8pt]
\begin{subtable}[t]{0.49\linewidth}\centering
\begin{tabular}{Lcccc}
\toprule
MoELoRA & OCR & Math & VP & APP \\
\midrule
OCR & 32.70 &  &  &  \\
Math & 32.50 & 33.30 &  &  \\
VP & 30.80 & 33.00 & 64.59 &  \\
APP & 26.70 & 32.80 & 56.85 & 27.22 \\
\bottomrule
\end{tabular}

\end{subtable}\hfill
\begin{subtable}[t]{0.49\linewidth}\centering
\begin{tabular}{Lcccc}
\toprule
MoELoRA$^*$ & OCR & Math & VP & APP \\
\midrule
OCR & 32.70 &  &  &  \\
Math & 29.40 & 33.10 &  &  \\
VP & 32.60 & 32.50 & 65.61 &  \\
APP & 19.80 & 32.20 & 54.19 & 30.00 \\
\bottomrule
\end{tabular}

\end{subtable}\\[8pt]
\begin{subtable}[t]{0.49\linewidth}\centering
\begin{tabular}{Lcccc}
\toprule
CL-MoE & OCR & Math & VP & APP \\
\midrule
OCR & 33.00 &  &  &  \\
Math & 32.30 & 33.60 &  &  \\
VP & 30.20 & 32.50 & 64.72 &  \\
APP & 19.90 & 32.70 & 53.43 & 30.69 \\
\bottomrule
\end{tabular}

\end{subtable}\hfill
\begin{subtable}[t]{0.49\linewidth}\centering
\begin{tabular}{Lcccc}
\toprule
CL-MoE$^*$ & OCR & Math & VP & APP \\
\midrule
OCR & 33.20 &  &  &  \\
Math & 34.30 & 36.70 &  &  \\
VP & 32.00 & 33.20 & 64.97 &  \\
APP & 25.40 & 31.80 & 60.91 & 30.00 \\
\bottomrule
\end{tabular}

\end{subtable}\\[8pt]
\begin{subtable}[t]{0.49\linewidth}\centering
\begin{tabular}{Lcccc}
\toprule
HiDe & OCR & Math & VP & APP \\
\midrule
OCR & 33.40 &  &  &  \\
Math & 30.90 & 32.80 &  &  \\
VP & 30.40 & 33.30 & 56.98 &  \\
APP & 24.60 & 32.10 & 46.32 & 28.75 \\
\bottomrule
\end{tabular}

\end{subtable}\hfill
\begin{subtable}[t]{0.49\linewidth}\centering
\begin{tabular}{Lcccc}
\toprule
HiDe$^*$ & OCR & Math & VP & APP \\
\midrule
OCR & 34.10 &  &  &  \\
Math & 32.60 & 35.70 &  &  \\
VP & 30.70 & 32.60 & 53.81 &  \\
APP & 24.60 & 28.40 & 30.71 & 23.75 \\
\bottomrule
\end{tabular}

\end{subtable}\\[8pt]
\begin{subtable}[t]{0.49\linewidth}\centering
\begin{tabular}{Lcccc}
\toprule
SEFE & OCR & Math & VP & APP \\
\midrule
OCR & 33.00 &  &  &  \\
Math & 32.20 & 32.60 &  &  \\
VP & 31.80 & 33.30 & 64.59 &  \\
APP & 26.00 & 33.40 & 57.74 & 33.75 \\
\bottomrule
\end{tabular}

\end{subtable}\hfill
\begin{subtable}[t]{0.49\linewidth}\centering
\begin{tabular}{Lcccc}
\toprule
SEFE$^*$ & OCR & Math & VP & APP \\
\midrule
OCR & 33.60 &  &  &  \\
Math & 33.80 & 37.50 &  &  \\
VP & 32.80 & 36.10 & 66.50 &  \\
APP & 25.60 & 34.80 & 57.61 & 31.39 \\
\bottomrule
\end{tabular}

\end{subtable}\\[8pt]
\begin{subtable}[t]{0.49\linewidth}\centering
\begin{tabular}{Lcccc}
\toprule
DISCO & OCR & Math & VP & APP \\
\midrule
OCR & 32.90 &  &  &  \\
Math & 31.80 & 33.40 &  &  \\
VP & 31.00 & 34.50 & 59.64 &  \\
APP & 32.90 & 33.10 & 60.15 & 30.14 \\
\bottomrule
\end{tabular}

\end{subtable}\hfill
\begin{subtable}[t]{0.49\linewidth}\centering
\begin{tabular}{Lcccc}
\toprule
DISCO$^*$ & OCR & Math & VP & APP \\
\midrule
OCR & 33.40 &  &  &  \\
Math & 32.10 & 36.60 &  &  \\
VP & 32.20 & 37.00 & 63.07 &  \\
APP & 34.20 & 35.00 & 61.55 & 27.50 \\
\bottomrule
\end{tabular}

\end{subtable}\\[8pt]

\end{table}

\newpage
\begin{table}[H]
\centering \setlength{\tabcolsep}{4pt} \small
\caption{Result matrices of InternVL-based baselines in ability continual learning. $^{*}$ denotes
the original method with replay data.}
\label{tab:aclresults:internvl}

\begin{subtable}[t]{0.49\linewidth}\centering
\begin{tabular}{Lcccc}
\toprule
LoRA-FT & OCR & Math & VP & APP \\
\midrule
OCR & 32.20 &  &  &  \\
Math & 33.10 & 33.30 &  &  \\
VP & 31.80 & 32.30 & 68.02 &  \\
APP & 21.40 & 32.80 & 60.28 & 29.86 \\
\bottomrule
\end{tabular}

\end{subtable}\hfill
\begin{subtable}[t]{0.49\linewidth}\centering
\begin{tabular}{Lcccc}
\toprule
LoRA-FT$^*$ & OCR & Math & VP & APP \\
\midrule
OCR & 31.60 &  &  &  \\
Math & 35.30 & 35.40 &  &  \\
VP & 32.60 & 31.10 & 68.27 &  \\
APP & 26.30 & 34.20 & 62.56 & 31.25 \\
\bottomrule
\end{tabular}

\end{subtable}\\[8pt]
\begin{subtable}[t]{0.49\linewidth}\centering
\begin{tabular}{Lcccc}
\toprule
O-LoRA & OCR & Math & VP & APP \\
\midrule
OCR & 32.70 &  &  &  \\
Math & 31.10 & 34.20 &  &  \\
VP & 30.20 & 33.00 & 63.20 &  \\
APP & 25.50 & 32.30 & 64.59 & 24.44 \\
\bottomrule
\end{tabular}

\end{subtable}\hfill
\begin{subtable}[t]{0.49\linewidth}\centering
\begin{tabular}{Lcccc}
\toprule
O-LoRA$^*$ & OCR & Math & VP & APP \\
\midrule
OCR & 34.00 &  &  &  \\
Math & 30.90 & 34.40 &  &  \\
VP & 31.00 & 33.20 & 65.86 &  \\
APP & 21.70 & 31.10 & 59.77 & 31.25 \\
\bottomrule
\end{tabular}

\end{subtable}\\[8pt]
\begin{subtable}[t]{0.49\linewidth}\centering
\begin{tabular}{Lcccc}
\toprule
MoELoRA & OCR & Math & VP & APP \\
\midrule
OCR & 32.20 &  &  &  \\
Math & 29.90 & 33.30 &  &  \\
VP & 29.20 & 32.80 & 67.64 &  \\
APP & 17.20 & 32.70 & 55.33 & 32.50 \\
\bottomrule
\end{tabular}

\end{subtable}\hfill
\begin{subtable}[t]{0.49\linewidth}\centering
\begin{tabular}{Lcccc}
\toprule
MoELoRA$^*$ & OCR & Math & VP & APP \\
\midrule
OCR & 32.90 &  &  &  \\
Math & 31.50 & 36.50 &  &  \\
VP & 30.90 & 32.30 & 65.74 &  \\
APP & 13.90 & 29.70 & 54.95 & 32.50 \\
\bottomrule
\end{tabular}

\end{subtable}\\[8pt]
\begin{subtable}[t]{0.49\linewidth}\centering
\begin{tabular}{Lcccc}
\toprule
HiDe & OCR & Math & VP & APP \\
\midrule
OCR & 33.40 &  &  &  \\
Math & 26.30 & 33.60 &  &  \\
VP & 30.10 & 33.00 & 61.80 &  \\
APP & 17.70 & 33.00 & 41.12 & 20.28 \\
\bottomrule
\end{tabular}

\end{subtable}\hfill
\begin{subtable}[t]{0.49\linewidth}\centering
\begin{tabular}{Lcccc}
\toprule
HiDe$^*$ & OCR & Math & VP & APP \\
\midrule
OCR & 33.40 &  &  &  \\
Math & 28.10 & 34.70 &  &  \\
VP & 31.10 & 32.20 & 55.33 &  \\
APP & 25.30 & 29.20 & 42.13 & 20.28 \\
\bottomrule
\end{tabular}

\end{subtable}\\[8pt]
\begin{subtable}[t]{0.49\linewidth}\centering
\begin{tabular}{Lcccc}
\toprule
DISCO & OCR & Math & VP & APP \\
\midrule
OCR & 31.90 &  &  &  \\
Math & 31.70 & 34.00 &  &  \\
VP & 32.10 & 33.50 & 63.45 &  \\
APP & 30.60 & 33.10 & 65.36 & 27.50 \\
\bottomrule
\end{tabular}

\end{subtable}\hfill
\begin{subtable}[t]{0.49\linewidth}\centering
\begin{tabular}{Lcccc}
\toprule
DISCO$^*$ & OCR & Math & VP & APP \\
\midrule
OCR & 34.70 &  &  &  \\
Math & 31.50 & 34.70 &  &  \\
VP & 31.50 & 34.60 & 62.31 &  \\
APP & 32.30 & 32.30 & 64.97 & 30.14 \\
\bottomrule
\end{tabular}

\end{subtable}\\[8pt]

\end{table}

\subsection{Detailed Results of~\ours}

\begin{table}[H]
\centering \setlength{\tabcolsep}{4pt} \small
\caption{Result matrices of~\ours~in domain continual learning. LLaVA denotes LLaVA-based MR-LoRA, and InternVL denotes InternVL-based MR-LoRA.}
\label{tab:mrlora:dcl}
\begin{subtable}[t]{0.49\linewidth}
\centering
\begin{tabular}{lccccc}
\toprule
LLaVA & RS & Med & AD & Sci & Fin \\
\midrule
RS & 81.06 &  &  &  &  \\
Med & 81.06 & 65.73 &  &  &  \\
AD & 81.06 & 65.71 & 54.17 &  &  \\
Sci & 81.06 & 65.68 & 54.17 & 56.11 &  \\
Fin & 80.87 & 65.32 & 54.12 & 56.71 & 91.12 \\
\bottomrule
\end{tabular}

\end{subtable}
\hfill
\begin{subtable}[t]{0.49\linewidth}
\centering
\begin{tabular}{lccccc}
\toprule
InternVL & RS & Med & AD & Sci & Fin \\
\midrule
RS & 81.49 &  &  &  &  \\
Med & 81.49 & 66.40 &  &  &  \\
AD & 81.49 & 66.42 & 54.56 &  &  \\
Sci & 81.47 & 65.81 & 54.56 & 54.05 &  \\
Fin & 81.48 & 65.80 & 54.56 & 54.40 & 91.07 \\
\bottomrule
\end{tabular}

\end{subtable}
\end{table}

\begin{table}[H]
\centering \setlength{\tabcolsep}{5pt} \small
\caption{Result matrices of~\ours~in ability continual learning. LLaVA denotes LLaVA-based MR-LoRA, and InternVL denotes InternVL-based MR-LoRA.}
\label{tab:mrlora:acl}
\begin{subtable}[t]{0.49\linewidth}
\centering
\begin{tabular}{lcccc}
\toprule
LLaVA & OCR & Math & VP & APP \\
\midrule
OCR & 33.60 &  &  &  \\
Math & 33.50 & 36.50 &  &  \\
VP & 33.50 & 36.40 & 64.97 &  \\
APP & 33.70 & 36.20 & 65.10 & 32.50 \\
\bottomrule
\end{tabular}

\end{subtable}
\hfill
\begin{subtable}[t]{0.49\linewidth}
\centering
\begin{tabular}{lcccc}
\toprule
InternVL & OCR & Math & VP & APP \\
\midrule
OCR & 32.20 &  &  &  \\
Math & 33.80 & 36.40 &  &  \\
VP & 33.30 & 35.60 & 67.89 &  \\
APP & 33.00 & 35.70 & 67.51 & 33.75 \\
\bottomrule
\end{tabular}

\end{subtable}
\end{table}

\clearpage
\section{Limitations and Broader Impacts}
\subsection{Limitations}\label{limitation}
Although our study makes valuable contributions, we acknowledge the following limitations:
(1)~Model size and training limitations: This research focuses exclusively on MLLMs with 7 billion parameters. Owing to computational constraints, we did not explore larger models. 
(2)~potential inaccuracies in the StockQA dataset:
Our StockQA dataset is generated by Qwen2.5-VL~\citep{Qwen2.5-VL}, and the model may inadvertently produce inaccurate or misleading data. 
Moreover, biases inherent in the training data could manifest in the generated dataset, influencing the outcomes and interpretations of subsequent analyses.
We hope to address these limitations in our future work to build a practical and lifelong-evolving MLLM.
\subsection{Broader Impacts}\label{broader}
 Positively, such work advances the ability of AI systems to learn adaptively from ongoing streams of diverse data, enabling applications in education, assistive technologies, and personalized healthcare. These systems could provide more context-aware and accessible tools that evolve over time to better support users' needs. Moreover, robust continual learning reduces the need for retraining from scratch, leading to more energy-efficient and sustainable AI development.
 However, there are potential negative impacts. Without careful design, continual learning systems may inadvertently retain or amplify biases from evolving data streams, leading to fairness concerns. The dynamic nature of these models also complicates auditing and accountability, as their behavior changes over time. Additionally, if misused, adaptive models could enhance surveillance or manipulation by continuously tailoring outputs to influence user behavior. To mitigate these risks, transparency, rigorous evaluation, and ethical safeguards must be integrated into both benchmark design and method development.

\section{Inference Optimization with Caching}
A key advantage of our method is its computational efficiency during inference. While our approach involves two distinct phases, we introduce a caching strategy that collapses the computational overhead. 
The most intensive operation—the forward pass through the backbone network (\textit{i.e.}, the visual encoder and LLM) is performed only once. 
We cache the resulting hidden states from each layer (specifically, the KV cache) after this single pass. 
Subsequently, our two lightweight modules, the router and the expert LoRA, operate sequentially on these cached states, obviating the need for a second full forward pass. 
This optimization reduces the computational cost from that of two full inferences to only marginally more than a single one, achieving a practical deployment cost comparable to standard single-pass methods, such as LoRA-FT~\citep{hu2021lora}.

\section{Use of LLM}
In the preparation of this manuscript, we utilized a Large Language Model (LLM) in a capacity analogous to a conventional grammar-checking tool. Its application was strictly confined to copy-editing tasks, such as correcting spelling, improving grammar, and enhancing the clarity of author-generated text. No part of the research ideation, methodology, data analysis, or generation of substantive content was performed by the LLM.

\clearpage

\section{Visualization}

\subsection{Illustration of MLLM-CL Benchmark} \label{appdx:bench:sample}
In this section, we show more examples of our MLLM-CL benchmark in domain continual learning and ability continual learning.
\begin{figure}[H]
    \centering
    \includegraphics[width=\linewidth]{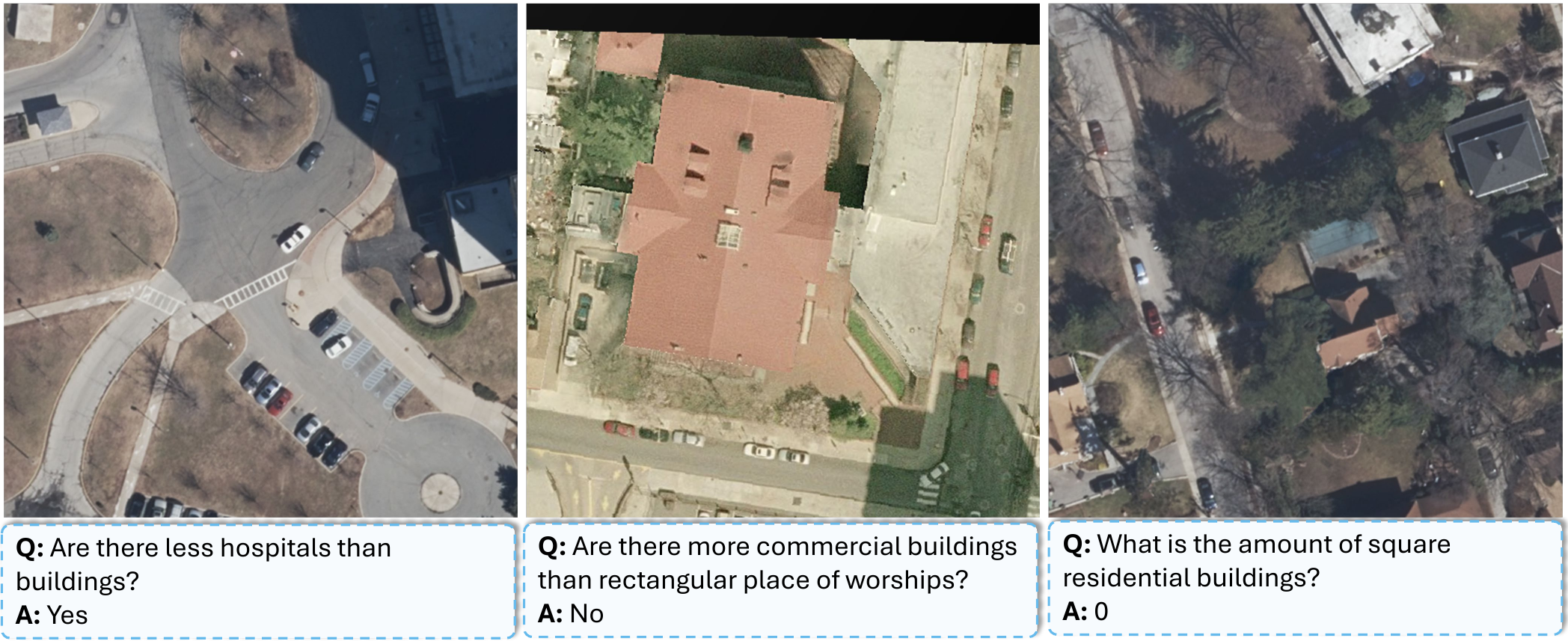}
    \caption{Examples of remote sensing task in domain continual learning.}
    \label{fig:rs}
\end{figure}
\begin{figure}[H]
    \centering
    \includegraphics[width=\linewidth]{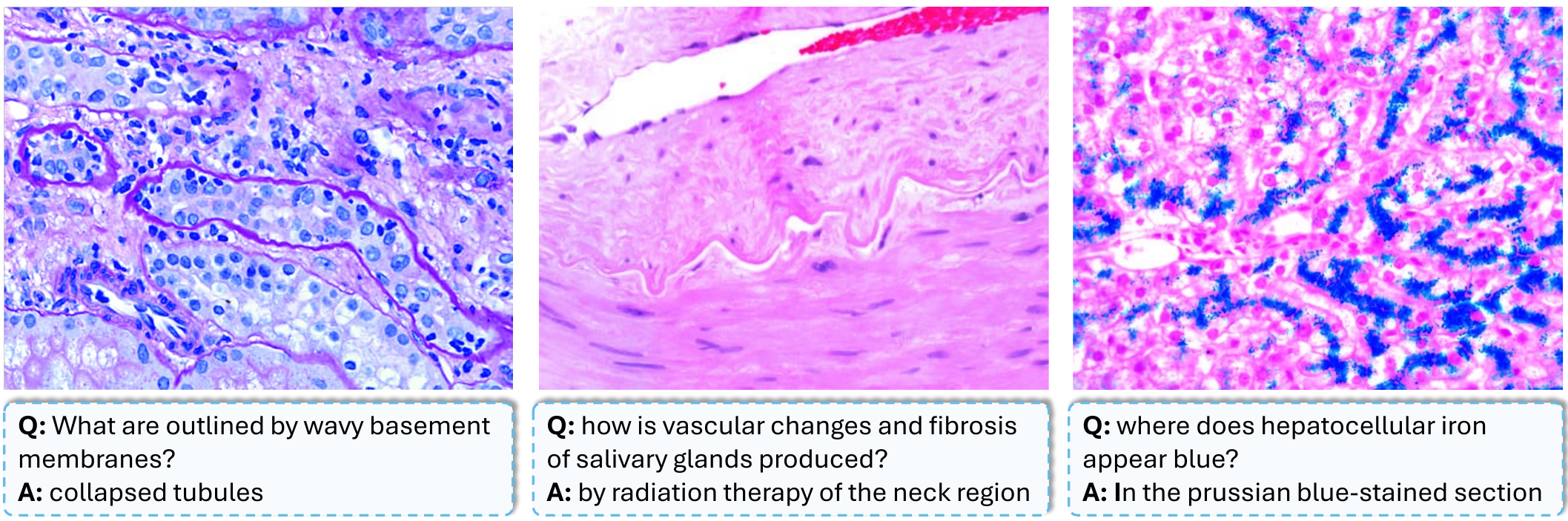}
    \caption{Examples of medical task in domain continual learning.}
\end{figure}
\begin{figure}[H]
    \centering
    \includegraphics[width=\linewidth]{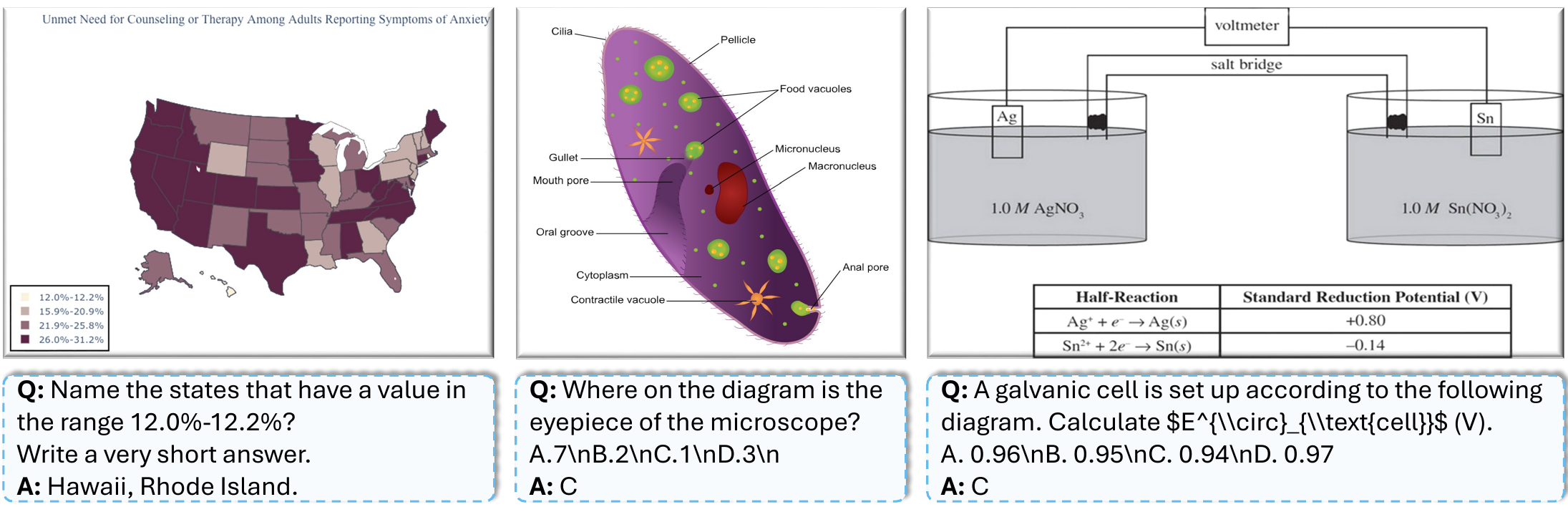}
    \caption{Examples of science task in domain continual learning.}
\end{figure}
\newpage
\begin{figure}[H]
    \centering
    \includegraphics[width=\linewidth]{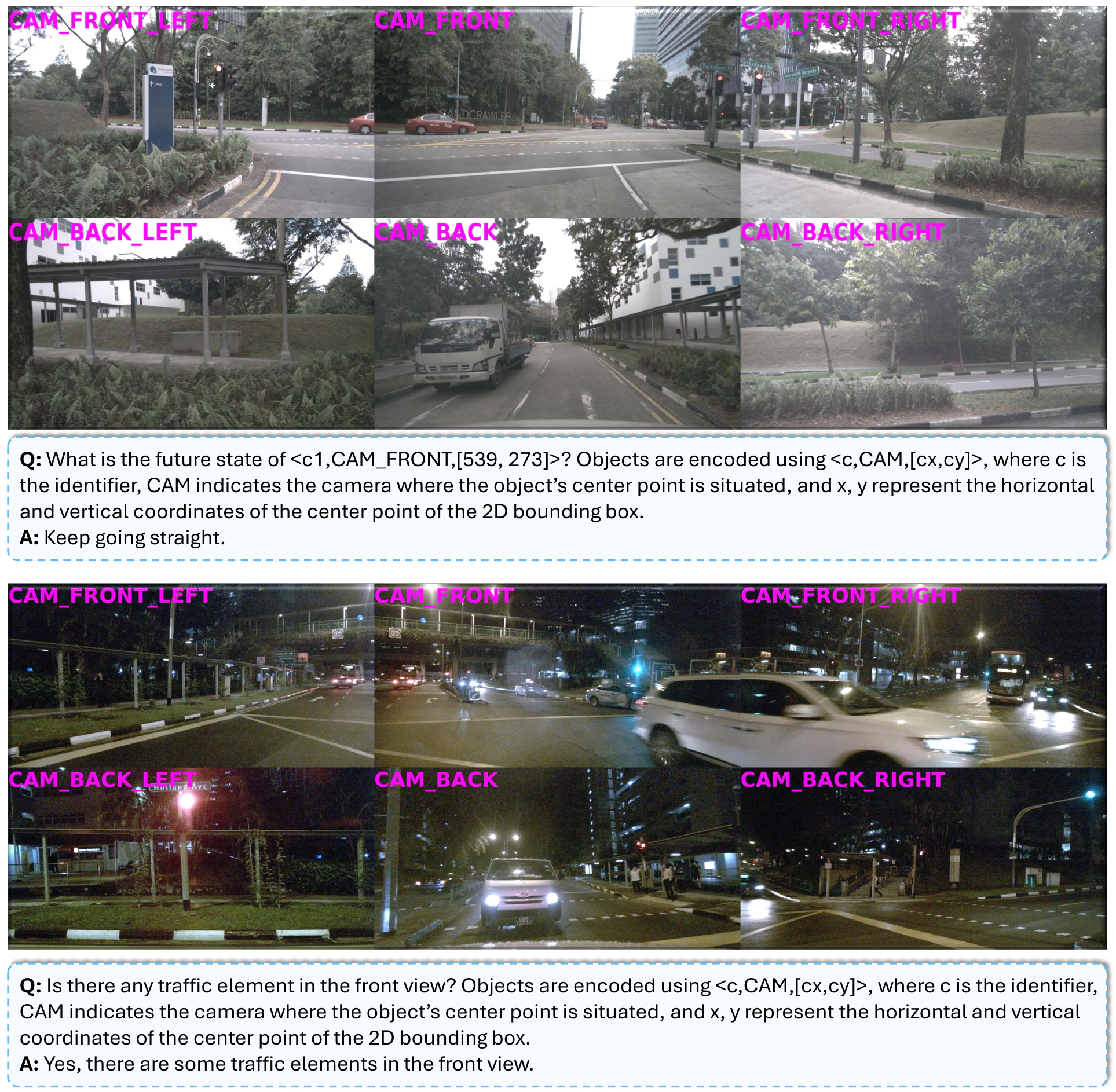}
    \caption{Examples of autonomous driving task in domain continual learning.}
\end{figure}
\begin{figure}[H]
    \centering
    \includegraphics[width=\linewidth]{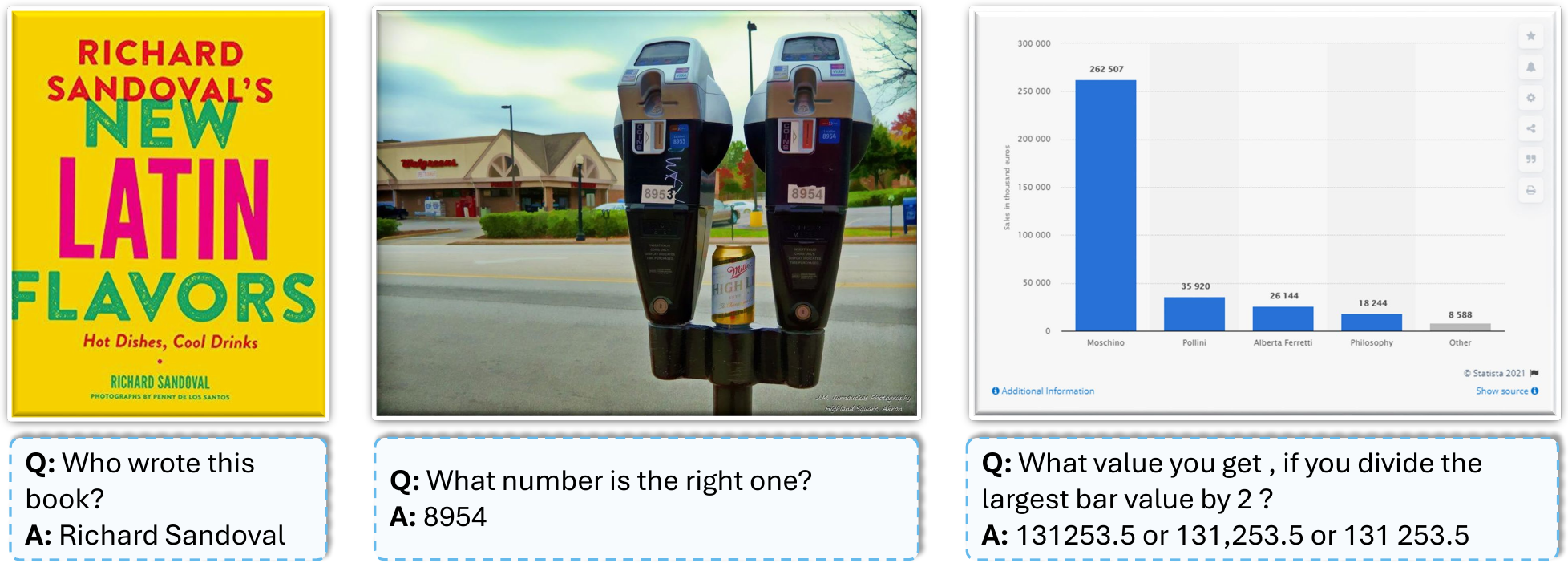}
    \caption{Examples of OCR task in ability continual learning.}
\end{figure}
\begin{figure}[H]
    \centering
    \includegraphics[width=\linewidth]{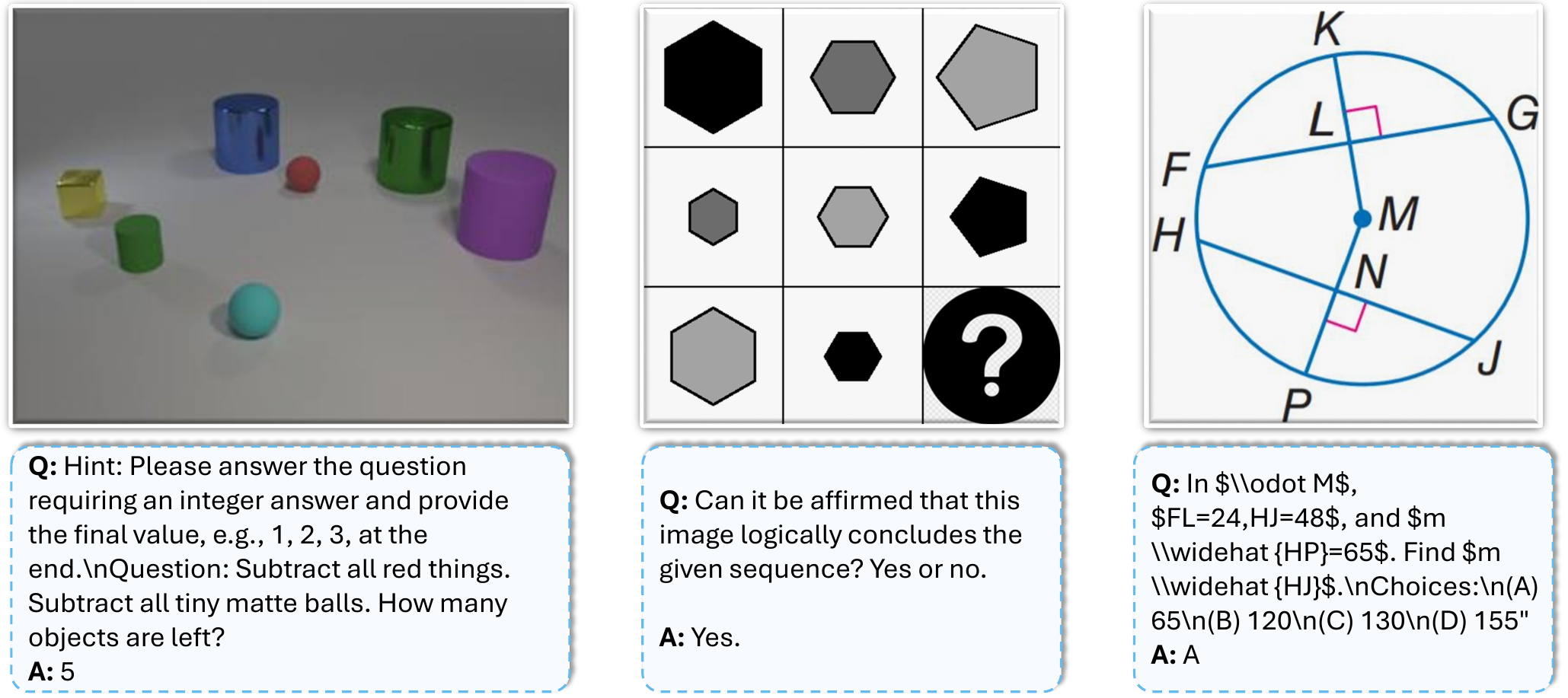}
    \caption{Examples of math task in ability continual learning.}
\end{figure}
\begin{figure}[H]
    \centering
    \includegraphics[width=\linewidth]{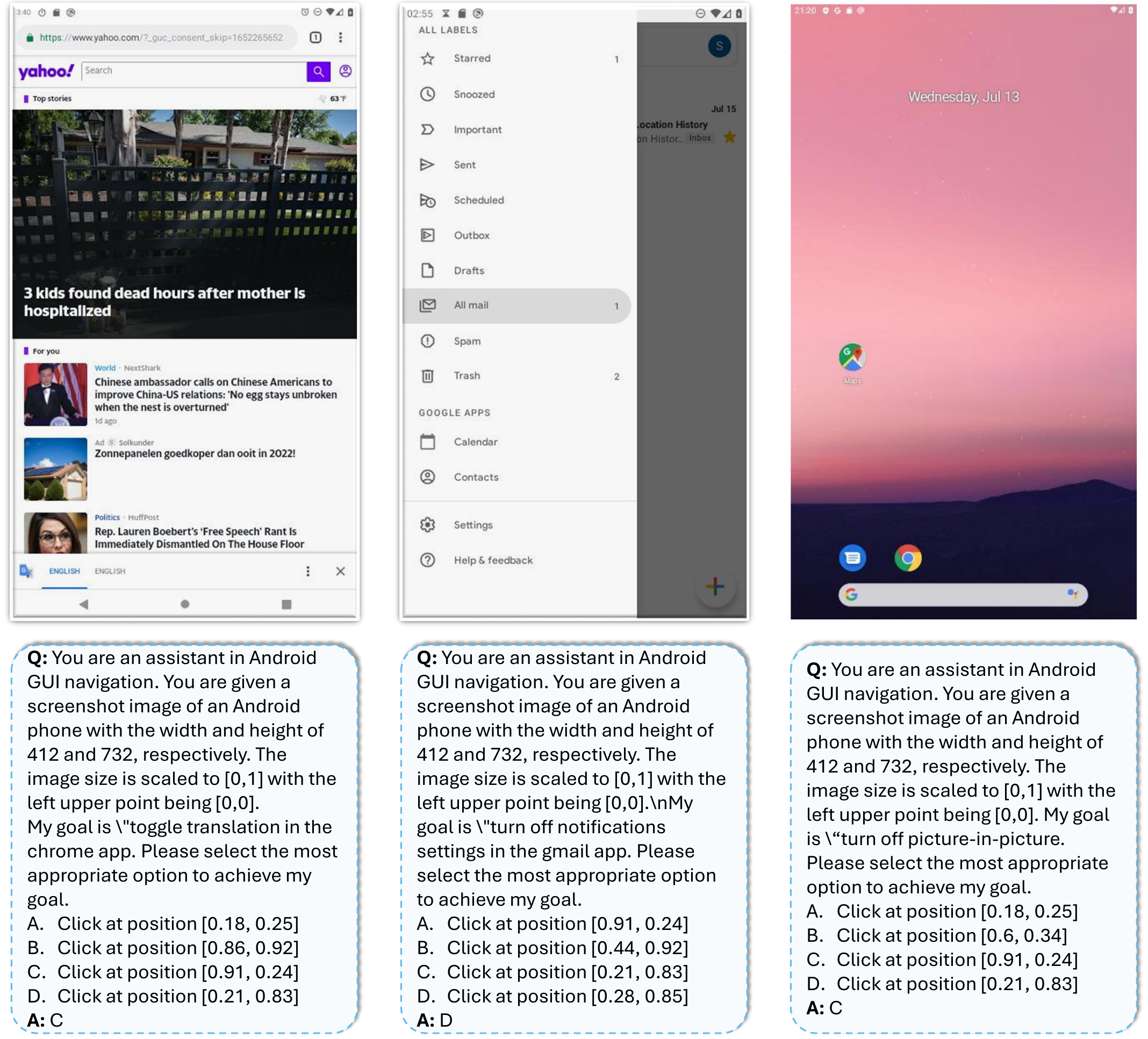}
    \caption{Examples of GUI agent task in ability continual learning.}
\end{figure}
\begin{figure}[H]
    \centering
    \includegraphics[width=\linewidth]{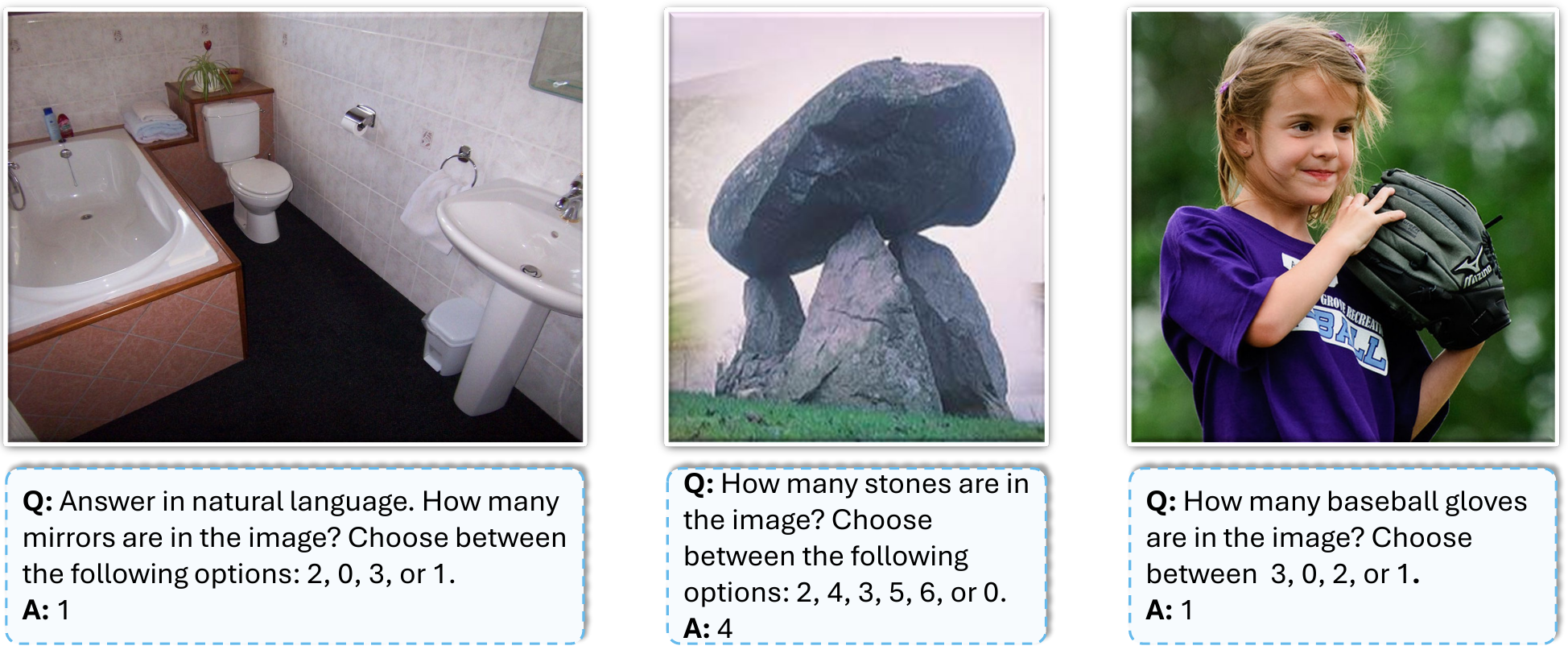}
    \caption{Examples of visual perception task in ability continual learning.}
\end{figure}

\subsection{Visualization of Results}
\begin{figure}[H]
    \centering
    \includegraphics[width=\linewidth]{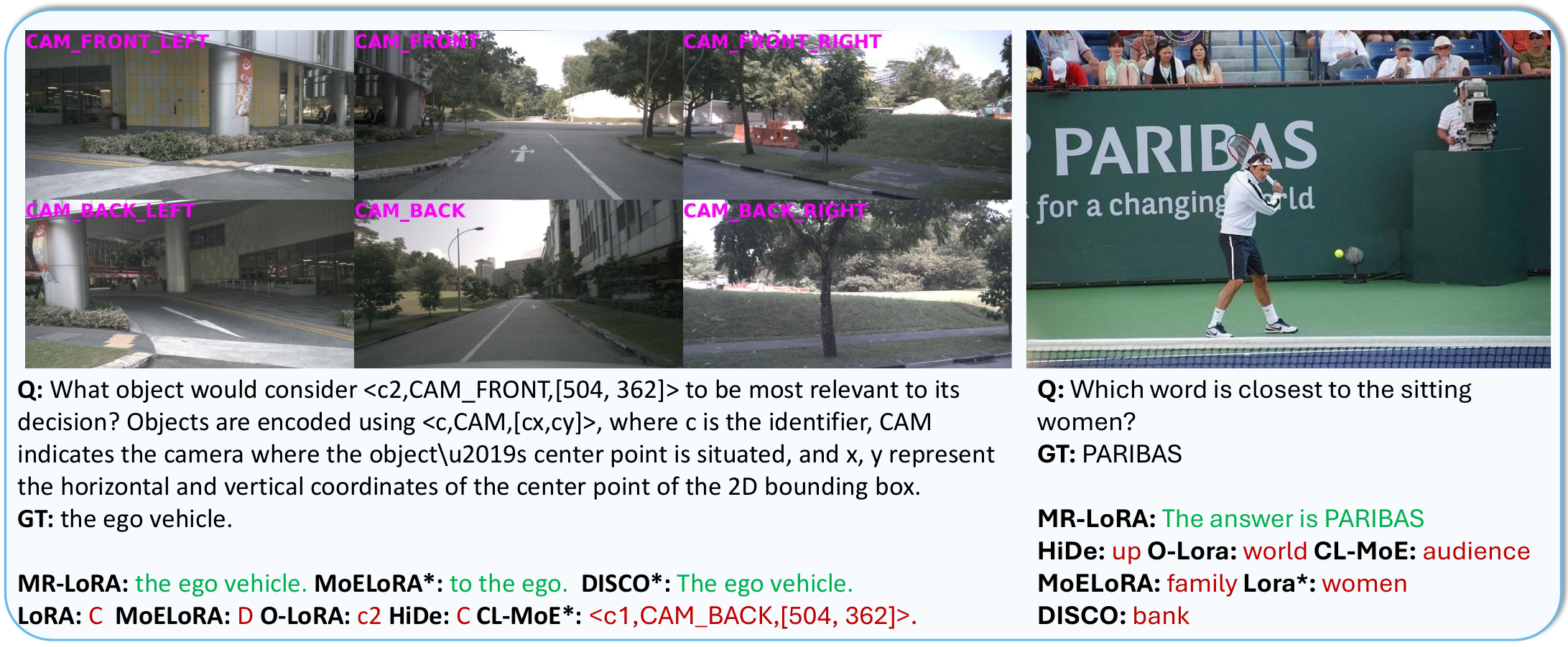}
    \vskip -0.05in
    \caption{Visualization of~\ours~and other baselines under domain continual learning and ability continual learning.
    The left part is testing the autonomous driving task after learning all domain tasks, while the right part is testing the OCR tasks after learning all ability tasks.}
    \label{fig:example}
\end{figure}

\Cref{fig:example} provides examples during DCL and ACL, respectively. 
We can find that some baselines like LoRA~\citep{hu2021lora}, MoELoRA~\citep{chen2024coin}, HiDE~\citep{guo2025hide} overfit to the last learned task and output options that do not exist in domain continual learning.
In ACL, most baselines, including HiDe~\citep{guo2025hide}, DISCO~\citep{guo2025federated}, CL-MoE~\citep{huai2025cl}, etc., miss part of their OCR ability and do not answer the question correctly.

\end{document}